\documentclass[10pt,twocolumn,letterpaper]{article}

\usepackage{cvpr}
\usepackage{times}
\usepackage{epsfig}
\usepackage{graphicx}
\usepackage{amsmath}
\usepackage{amssymb}
\usepackage{booktabs}
\usepackage[numbers,sort]{natbib}
\usepackage{xcolor, colortbl}
\usepackage{tabularx}
\usepackage{multirow}

\usepackage{url}

\usepackage{color}
\usepackage{multirow}
\usepackage{subfigure}
\usepackage{mwe}
\usepackage{booktabs,arydshln}
 \usepackage{bbm}

\usepackage{amsmath,amsfonts,bm}

\def\eqref#1{Eq.~(\ref{#1})}
\def\Eqref#1{Equation~\ref{#1}}

\def\1{\bm{1}}

\def\rvx{{\mathbf{x}}}

\def\rmI{{\mathbf{I}}}

\DeclareMathAlphabet{\mathsfit}{\encodingdefault}{\sfdefault}{m}{sl}
\SetMathAlphabet{\mathsfit}{bold}{\encodingdefault}{\sfdefault}{bx}{n}

\DeclareMathOperator*{\argmax}{arg\,max}

\newcolumntype{C}[1]{>{\centering\let\newline\\\arraybackslash\hspace{0pt}}p{#1}}

\def\ie{\emph{i.e.}}
\def\eg{\emph{e.g.}}

\definecolor{brown}{rgb}{0.65, 0.16, 0.16}
\definecolor{purp}{rgb}{0.65, 0.16, 0.65}

\usepackage[pagebackref=true,breaklinks=true,letterpaper=true,colorlinks,bookmarks=false]{hyperref}

 \cvprfinalcopy 

\ifcvprfinal\fi
\begin{document}

\title{Learning for Single-Shot Confidence Calibration in Deep Neural Networks \\ through Stochastic Inferences}

\author{
Seonguk Seo\footnotemark[1]\thanks{Equal contribution} ~$^1$ \qquad Paul Hongsuck Seo\footnotemark[1] ~$^{1, 2}$ \qquad Bohyung Han$^1$ \\
 $^1$Computer Vision Lab., ECE \& ASRI, Seoul National University, Korea\\
 $^2$Computer Vision Lab., POSTECH, Korea\\
 {\tt\small \{seonguk, bhhan\}@snu.ac.kr \quad hsseo@postech.ac.kr}
}

\maketitle


\begin{abstract}
We propose a generic framework to calibrate accuracy and confidence of a prediction in deep neural networks through stochastic inferences.
We interpret stochastic regularization using a Bayesian model, and analyze the relation between predictive uncertainty of networks and variance of the prediction scores obtained by stochastic inferences for a single example.
Our empirical study shows that the accuracy and the score of a prediction are highly correlated with the variance of multiple stochastic inferences given by stochastic depth or dropout.
Motivated by this observation, we design a novel variance-weighted confidence-integrated loss function that is composed of two cross-entropy loss terms with respect to ground-truth and uniform distribution, which are balanced by variance of stochastic prediction scores.
The proposed loss function enables us to learn deep neural networks that predict confidence calibrated scores using a single inference.
Our algorithm presents outstanding confidence calibration performance and improves classification accuracy when combined with two popular stochastic regularization techniques---stochastic depth and dropout---in multiple models and datasets; it alleviates overconfidence issue in deep neural networks significantly by training networks to achieve prediction accuracy proportional to confidence of prediction.
\end{abstract}


\section{Introduction}

\begin{figure*}
\centering
       \subfigure[Baseline (ECE = 0.346)]{\includegraphics[width=0.307\linewidth]{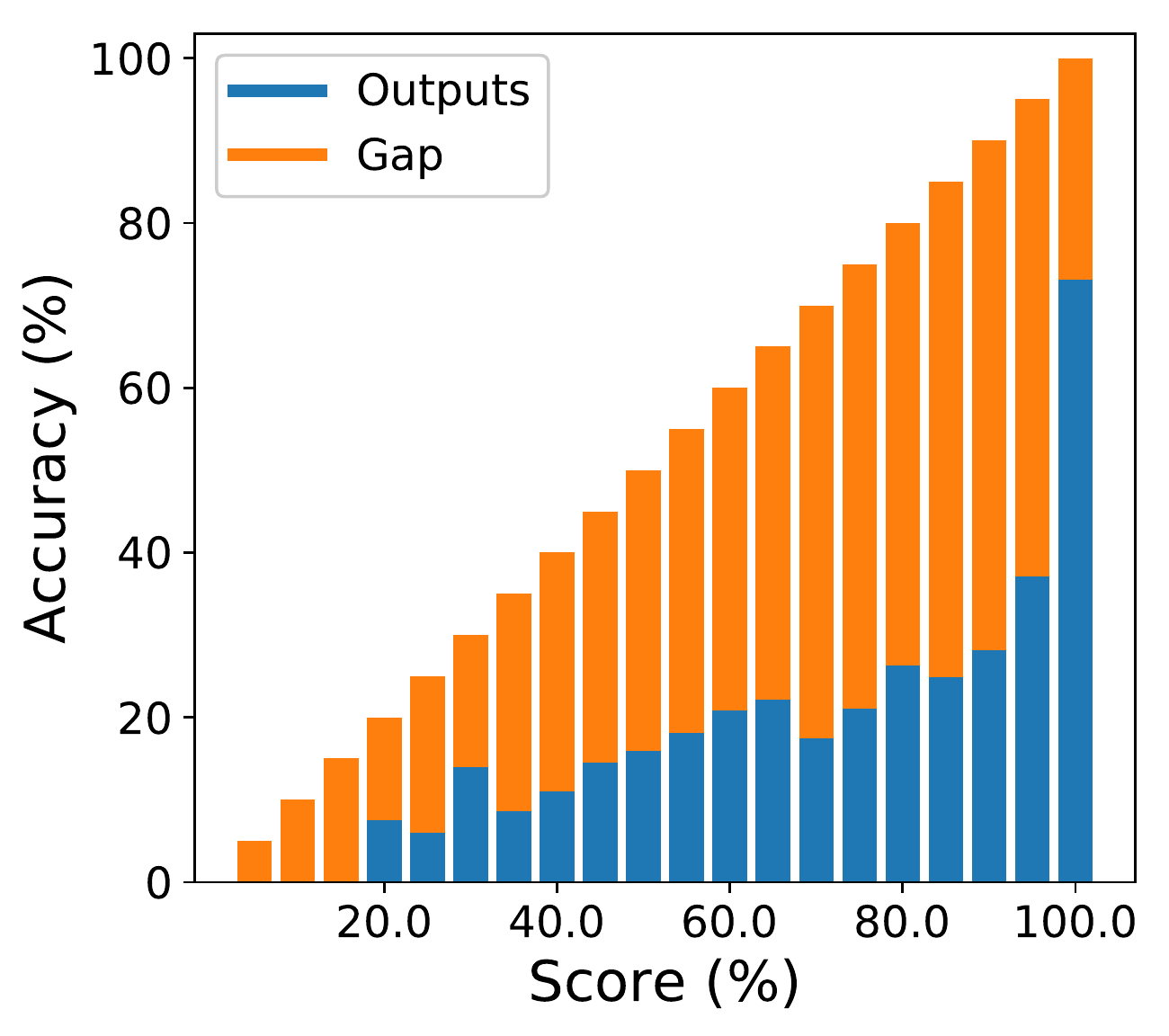}}
	\subfigure[CI{[Oracle]} (ECE = 0.122)]{\includegraphics[width=0.29\linewidth]{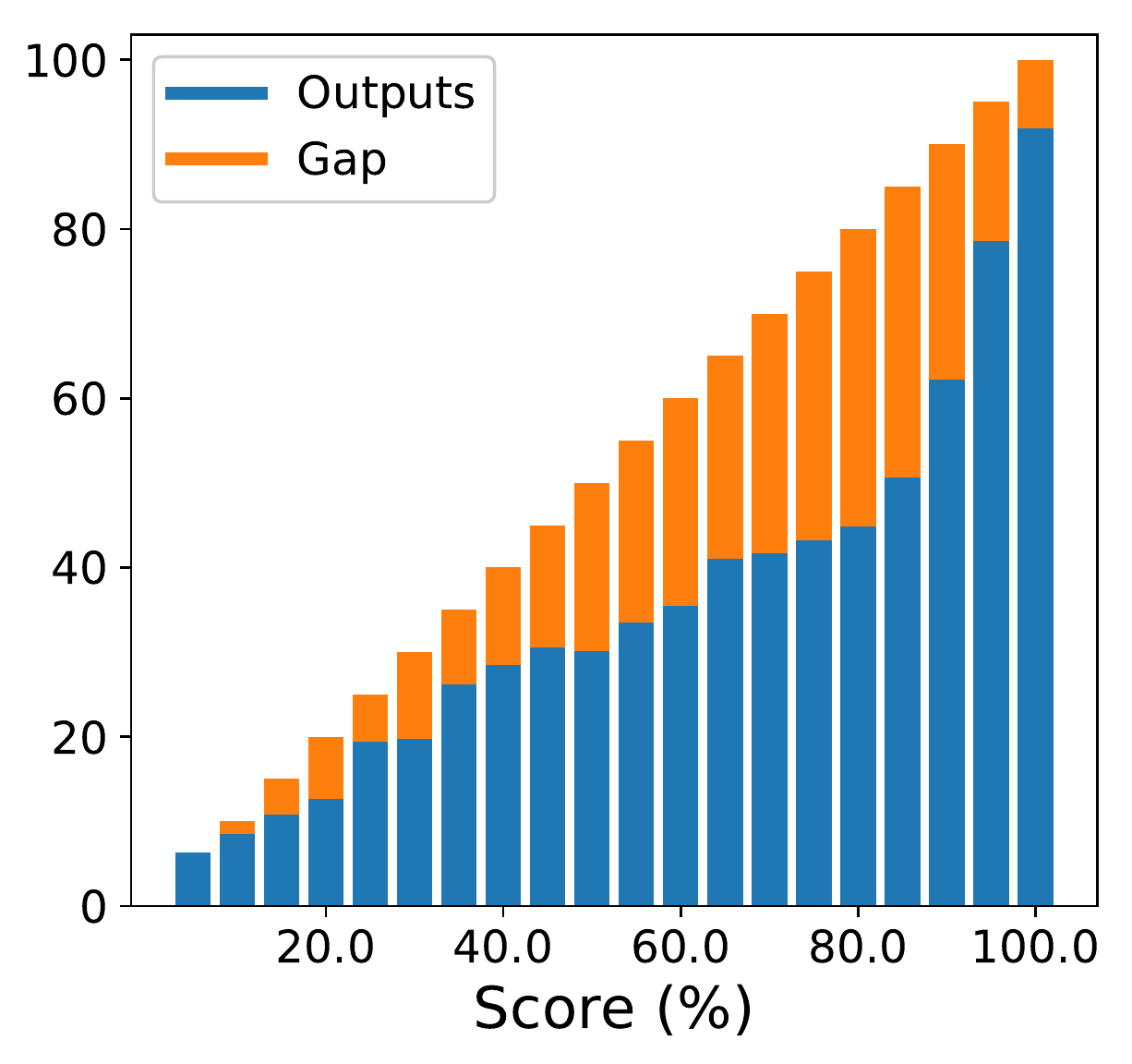}}
	 \subfigure[VWCI (ECE = 0.053)]{\includegraphics[width=0.29\linewidth]{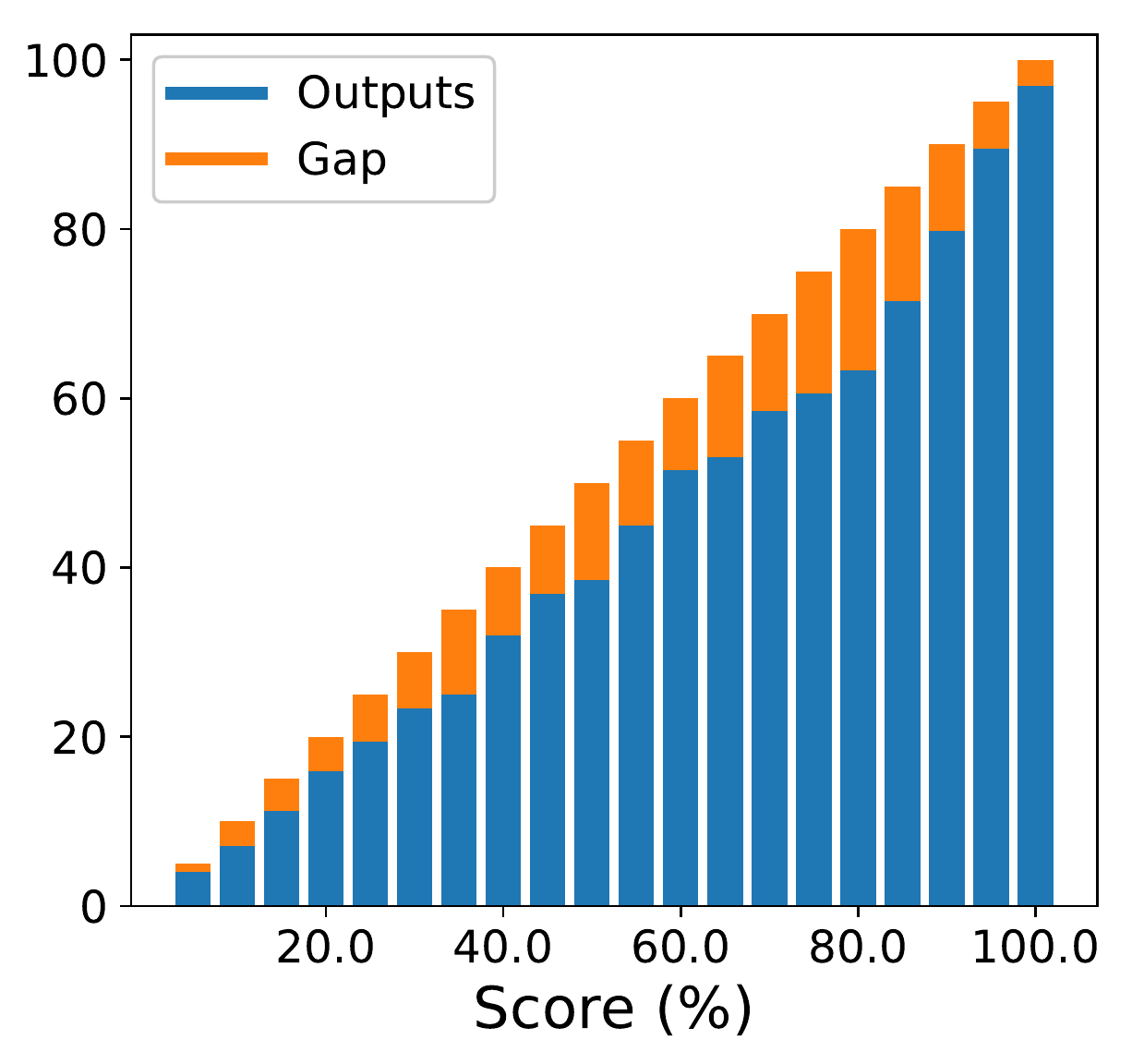}}
   \caption{Reliability diagrams of VGG-16 models trained with baseline, CI (ours) and VWCI (ours) losses in Tiny ImageNet dataset. This diagram shows expected accuracy as a function of confidence, \ie, classification score. ECE (Expected Calibration Error) denotes the average gap between confidence and expected accuracy. The proposed algorithm (VWCI) achieves well-calibrated results compared to the baseline and the best estimate by a simpler version of ours (CI).}
    \vspace{-0.2cm}
    \label{fig:var_acc_hist}
\end{figure*}

Deep neural networks have achieved remarkable performance in various tasks, but have critical limitations in reliability of their predictions.
One example is that inference results are often overly confident even for unseen or ambiguous examples. 
Since many practical applications including medical diagnosis, autonomous driving, and machine inspection require accurate uncertainty estimation as well as high prediction score for each inference, such an overconfidence issue makes deep neural networks inappropriate to be deployed for real-world problems in spite of their impressive accuracy.

Regularization is a common technique in training deep neural networks to avoid overfitting problems and improve generalization performance~\citep{srivastava2014dropout,huang2016deep,ioffe2015batch}.
Although regularization is effective to learn robust models, its objective is not directly related to generating score distributions aligned with uncertainty of predictions.
Hence, existing deep neural networks are often poor at calibrating prediction accuracy and confidence.

Our goal is to learn deep neural networks that are able to estimate uncertainty of each prediction while maintaining accuracy.
In other words, we propose a generic framework to calibrate prediction score (confidence) with accuracy in deep neural networks.
The main idea of our algorithm starts with an observation that the variance of prediction scores measured from multiple stochastic inferences is highly correlated with the accuracy and confidence of the average prediction.
We also show that a Bayesian interpretation of stochastic regularizations such as stochastic depth and dropout leads to the consistent conclusion with the observation. 
By using the empirical observation with the theoretical interpretation, we design a novel loss function to enable a deep neural network to predict confidence-calibrated scores based only on a single prediction, without multiple stochastic inferences.
Our contribution is summarized as
\begin{itemize} 
    \item We provide a generic framework to estimate uncertainty of a prediction based on stochastic inferences in deep neural networks, which is supported by empirical observations and theoretical analysis.
\vspace{-0.2cm}
    \item We propose a novel variance-weighted confidence-integrated loss function in a principled way, which enables networks to produce confidence-calibrated predictions even without performing stochastic inferences and introducing hyper-parameters.
\vspace{-0.2cm}

    \item The proposed framework presents outstanding performance to reduce overconfidence issue and estimate accurate uncertainty in various combinations of network architectures and datasets.
\end{itemize}

The rest of the paper is organized as follows.
We review the prior research and describe the theoretical background in Section~\ref{sec:related} and \ref{sec:preliminaries}, respectively.
Section~\ref{sec:confidence} presents our confidence calibration algorithm through stochastic inferences, and Section~\ref{sec:experiments} demonstrates experimental results.


\section{Related Work}
\label{sec:related}

Uncertainty modeling and estimation in deep neural networks is a critical problem and receives growing attention from machine learning community.
Bayesian approach is a common tool to provide a mathematical framework for uncertainty estimation.
However, the exact Bayesian inference is not tractable in deep neural networks due to its high computational cost, and various approximate inference techniques---MCMC~\citep{neal1996bayesian}, Laplace approximation~\citep{mackay1992bayesian} and variational inference~\citep{barber1998ensemble, graves2011varinf, hoffman2013stocvarinf, pawlowski2017implicit}---have been proposed. 
Recently, a Bayesian interpretation of multiplicative noise is employed to estimate uncertainty in deep neural networks~\citep{gal2016dropout, patrick2016representation}. 
Besides, there are several approaches outside Bayesian modeling, \eg, post-processing~\citep{niculescu2005predicting, john2000platt,zadrozny2001calibration, guo2017oncalibration} and deep ensembles~\citep{balaji2017simple}.
All the post-processing methods require a hold-out validation set to adjust prediction scores after training, and the ensemble-based technique employs multiple models to estimate uncertainty. 

Stochastic regularization is a well-known technique to improve generalization performance by injecting random noise to deep neural networks. 
The most notable method is dropout~\cite{srivastava2014dropout}, which rejects a subset of hidden units in a layer based on Bernoulli random noise. 
There exist several variants, for example, dropping weights~\citep{wan2013regularization} or skipping layers~\citep{huang2016deep}. 
Most stochastic regularization methods perform stochastic inferences during training, but make deterministic predictions using the full network during testing. 
On the contrary, we also employ stochastic inferences to obtain diverse and reliable outputs during testing.

Although the following works do not address uncertainty estimation, their main idea is related to our objective more or less.
Label smoothing~\citep{szegedy2015rethinking} encourages models to be less confident, by preventing a network from assigning the full probability to a single class. 
A similar loss function is discussed to train confidence-calibrated classifiers in \cite{lee2018training}, but it focuses on how to discriminate in-distribution and out-of-distribution examples, rather than estimating uncertainty or alleviating miscalibration of in-distribution examples.
On the other hand, \cite{pereyra2017regularizing} claims that blind label smoothing and penalizing entropy enhances accuracy by integrating loss functions with the same concept with \cite{szegedy2015rethinking, lee2018training}, but its improvement is marginal in practice.

\section{Preliminaries}
\label{sec:preliminaries}

This section describes a Bayesian interpretation of stochastic regularization in deep neural networks, and discusses the relationship between stochastic regularization and uncertainty modeling.

\subsection{Stochastic Methods for Regularizations}
A popular class of regularization techniques is stochastic regularization, which introduces random noise for perturbing network structures.
Our approach focuses on the multiplicative binary noise injection, where random binary noise is applied to the inputs or weights by elementwise multiplication, since such stochastic regularization techniques are widely used~\citep{srivastava2014dropout,wan2013regularization,huang2016deep}.
Note that input perturbation can be reformulated as weight perturbation.
For example, dropout---binary noise injection to activations---is interpretable as weight perturbation that masks out all the weights associated with the dropped inputs.
Therefore, if a classification network modeling $p(y|x,\theta)$ with parameters $\theta$ is trained with stochastic regularization methods by minimizing cross entropy, the loss function is defined by
\begin{equation}
    \mathcal{L}_\mathrm{SR}(\theta)= -\frac{1}{N}\sum_{i=1}^{N}{\log p\left(y_i\middle|x_i,\hat{\omega}_i\right)},
    \label{eq:srt}
\end{equation}
where $\hat{\omega}_{i}=\theta\odot\epsilon_i$ is a set of perturbed parameters by elementwise multiplication with random noise sample $\epsilon_i \sim p(\epsilon)$, and $(x_i,y_i) \in \mathcal{D}$ is a pair of input and output in training dataset $\mathcal{D}$.

At inference time, the network is parameterized by the expectation of the perturbed parameters, $\Theta=\mathbb{E}[\omega]=\theta\odot\mathbb{E}[\epsilon]$, to predict an output $\hat{y}$, which is given by
\begin{equation}
    \hat{y} = \argmax_y p\left(y|x,\Theta\right).
    \label{eq:srt_infer}
\end{equation}

\subsection{Bayesian Modeling}
Given the dataset $\mathcal{D}$ with $N$ examples, Bayesian objective is to estimate the posterior distribution of the model parameter, denoted by $p(\omega|\mathcal{D})$, to predict a label $y$ for an input $x$, which is given by
\begin{equation}
    p(y|x,\mathcal{D}) = \int_\omega p(y|x, \omega)p(\omega|\mathcal{D})d\omega.
\end{equation}
A common technique for the posterior estimation is variational approximation, which introduces an approximate distribution $q_\theta(\omega)$ and minimizes Kullback-Leibler (KL) divergence with the true posterior $D_\mathrm{KL}(q_\theta(\omega)||p(\omega|D))$ as follows:
\begin{align}
    \mathcal{L}_\mathrm{VA}(\theta)= &-\sum_{i=1}^N\int_\omega q_\theta(\omega)\log{p(y_i|x_i,\omega)}d\omega \nonumber \\&+ D_\mathrm{KL}(q_\theta(\omega)||p(\omega)).
    \label{eq:elbo}
\end{align}
The intractable integration and summation over the entire dataset in \eqref{eq:elbo} is approximated by Monte Carlo method and mini-batch optimization, resulting in
\begin{align}
    \hat{\mathcal{L}}_\mathrm{VA}(\theta)=&-\frac{N}{MS}\sum_{i=1}^M\sum_{j=1}^S\log{p\left(y_i\middle|x_i,\hat{\omega}_{i,j}\right)} \nonumber \\&+ D_\mathrm{KL}\left(q_\theta(\omega)\middle||p(\omega)\right),
    \label{eq:elbo_mc}
\end{align}
where $\hat{\omega}_{i,j}\sim q_\theta(\omega)$ is a sample from the approximate distribution, $S$ is the number of samples, and $M$ is the size of a mini-batch.
Note that the first term is data likelihood and the second term is divergence of the approximate distribution with respect to the prior distribution.

\subsection{Bayesian View of Stochastic Regularization}
\label{sec:sr_as_bayesian}
Suppose that we train a classifier with $\ell_2$ regularization by a stochastic gradient descent method.
Then, the loss function in \eqref{eq:srt} is rewritten as
\begin{equation}
    \hat{\mathcal{L}}_\mathrm{SR}(\theta)= -\frac{1}{M}\sum_{i=1}^{M}{\log p\left(y_i\middle|x_i,\hat{\omega}_i\right)} + \lambda ||\theta||_2^2,
    \label{eq:srt_wd}
\end{equation}
where $\ell_2$ regularization is applied to the deterministic parameters $\theta$ with weight $\lambda$.
Optimizing this loss function is equivalent to optimizing \eqref{eq:elbo_mc} if there exists a proper prior $p(\omega)$ and $q_\theta(\omega)$ is approximated as a Gaussian mixture distribution~\citep{gal2016dropout}.
Note that \cite{gal2016dropout} casts dropout training as an approximate Bayesian inference.
Thus, we can interpret training with stochastic depth~\citep{huang2016deep} within the same framework by a simple modification.
(See our supplementary document for the details.)
Then, the predictive distribution of a model trained with stochastic regularization is approximately given by
\begin{equation}
    \hat{p}(y|x,\mathcal{D}) = \int_\omega p(y|x,\omega) q_\theta(\omega) d\omega.
\end{equation}
Following \cite{gal2016dropout} and \cite{teye2018bayesian}, we estimate the predictive mean and uncertainty using a Monte Carlo approximation by drawing parameter samples $\{ \hat{\omega}_i \}_{i=1}^T$ as
\begin{align}
    \mathbb{E}_{\hat{p}}[y=c] \approx \frac{1}{T}\sum_{i=1}^T \hat{p}(y=c|x,\hat{\omega}_i), \label{eq:mean}\\
    \mathrm{Cov}_{\hat{p}}[\mathbf{y}] \approx  \mathbb{E}_{\hat{p}}[\mathbf{y} \mathbf{y}^\intercal] - \mathbb{E}_{\hat{p}}[\mathbf{y}] \mathbb{E}_{\hat{p}}[\mathbf{y}]^\intercal,
    \label{eq:variance}
\end{align}
where $\mathbf{y} = (y_1, \dots, y_C)^\intercal$ denotes a score vector of $C$ class labels.
\eqref{eq:mean} and \eqref{eq:variance} mean that the average prediction and its predictive uncertainty can be estimated from multiple stochastic inferences.


\section{Methods}
\label{sec:confidence}
We present a novel confidence calibration technique for prediction in deep neural networks, which is given by a variance-weighted confidence-integrated loss function.
We present our observation that variance of multiple stochastic inferences is closely related to accuracy and confidence of predictions, and provide an end-to-end training framework for confidence self-calibration.
Then, we show that the prediction accuracy and uncertainty are directly accessible from a predicted score from a single forward pass.

\begin{figure*}
    \centering
    \subfigure[Prediction uncertainty characteristics with stochastic depth in ResNet-34]{
    	\includegraphics[width=0.33\linewidth]{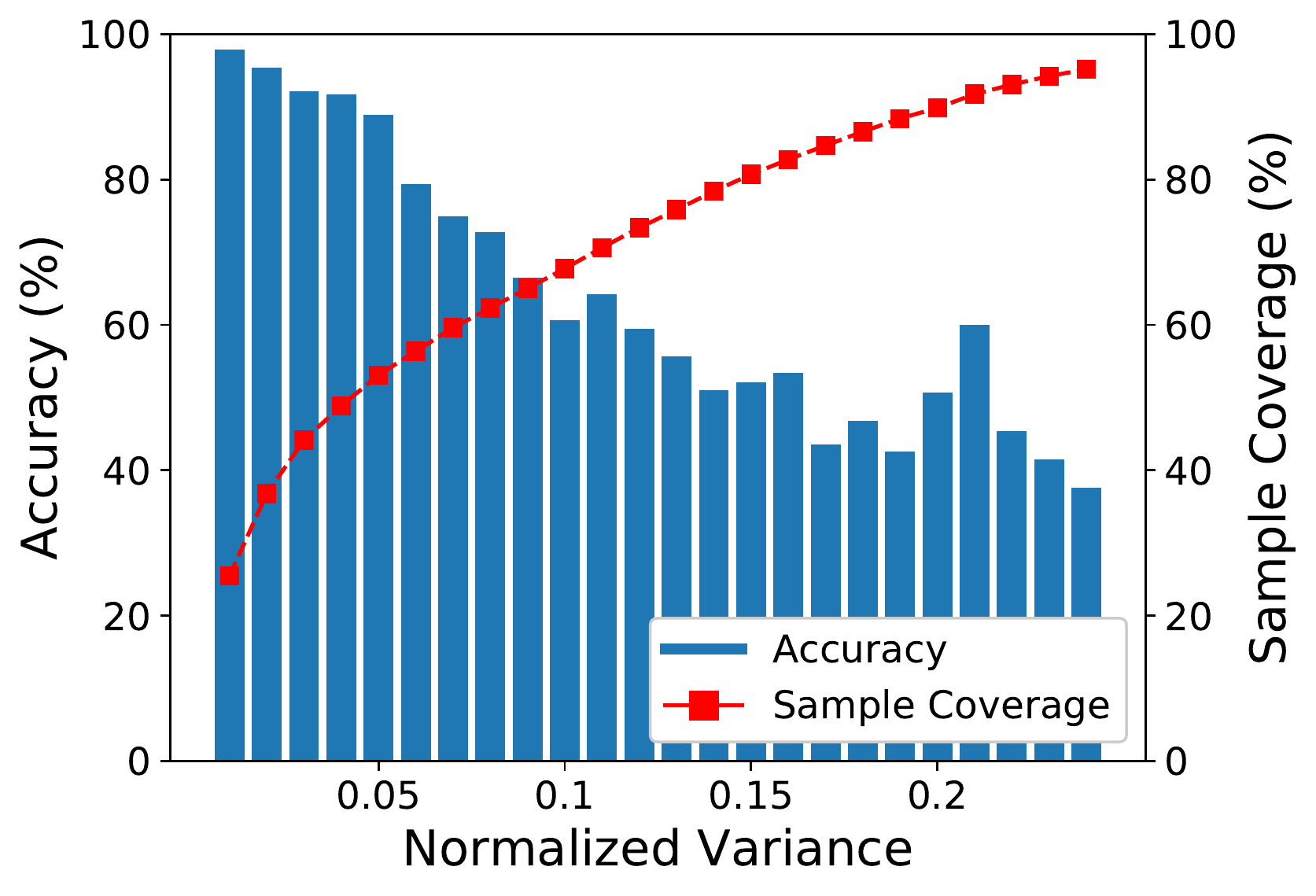}
    	\includegraphics[width=0.33\linewidth]{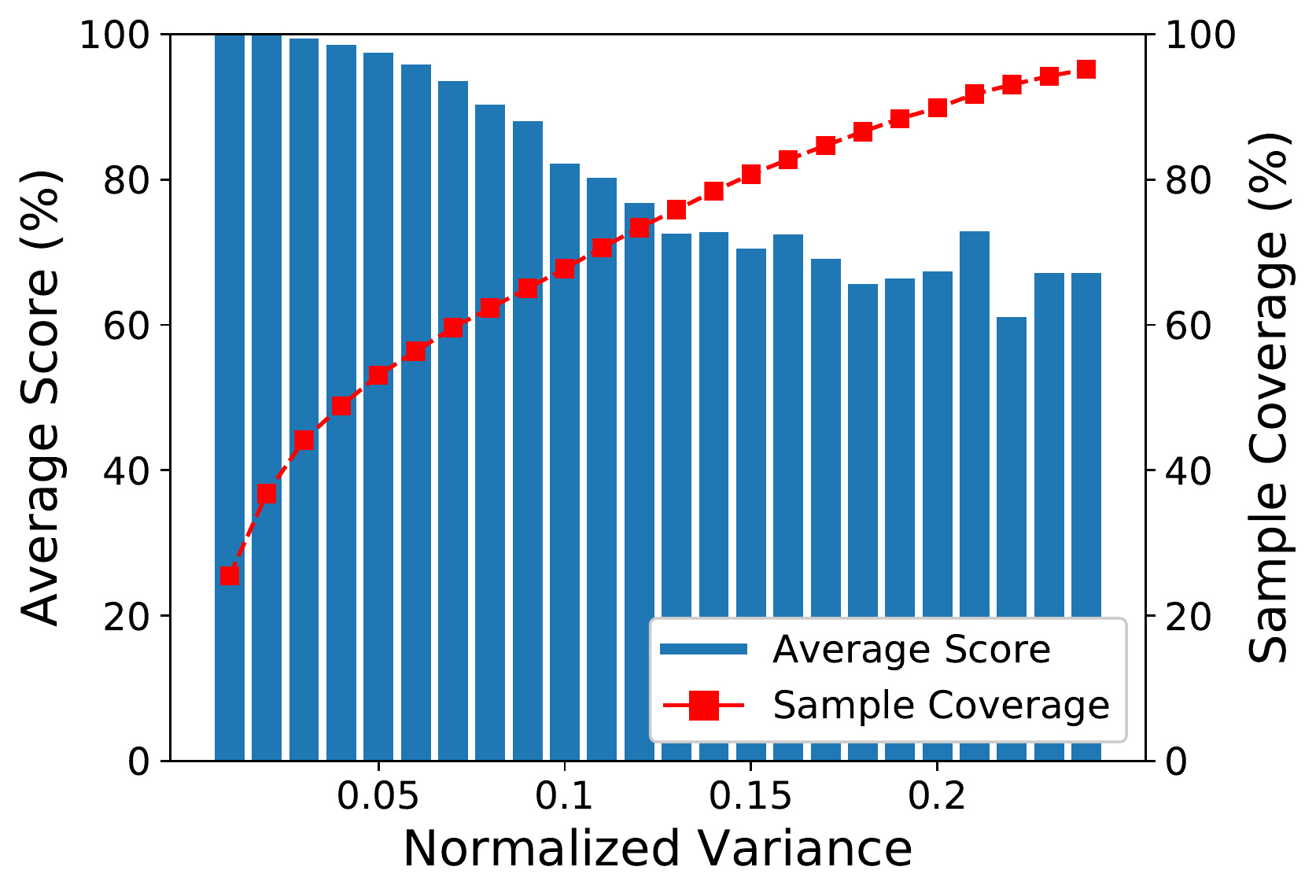}
    	\includegraphics[width=0.295\linewidth]{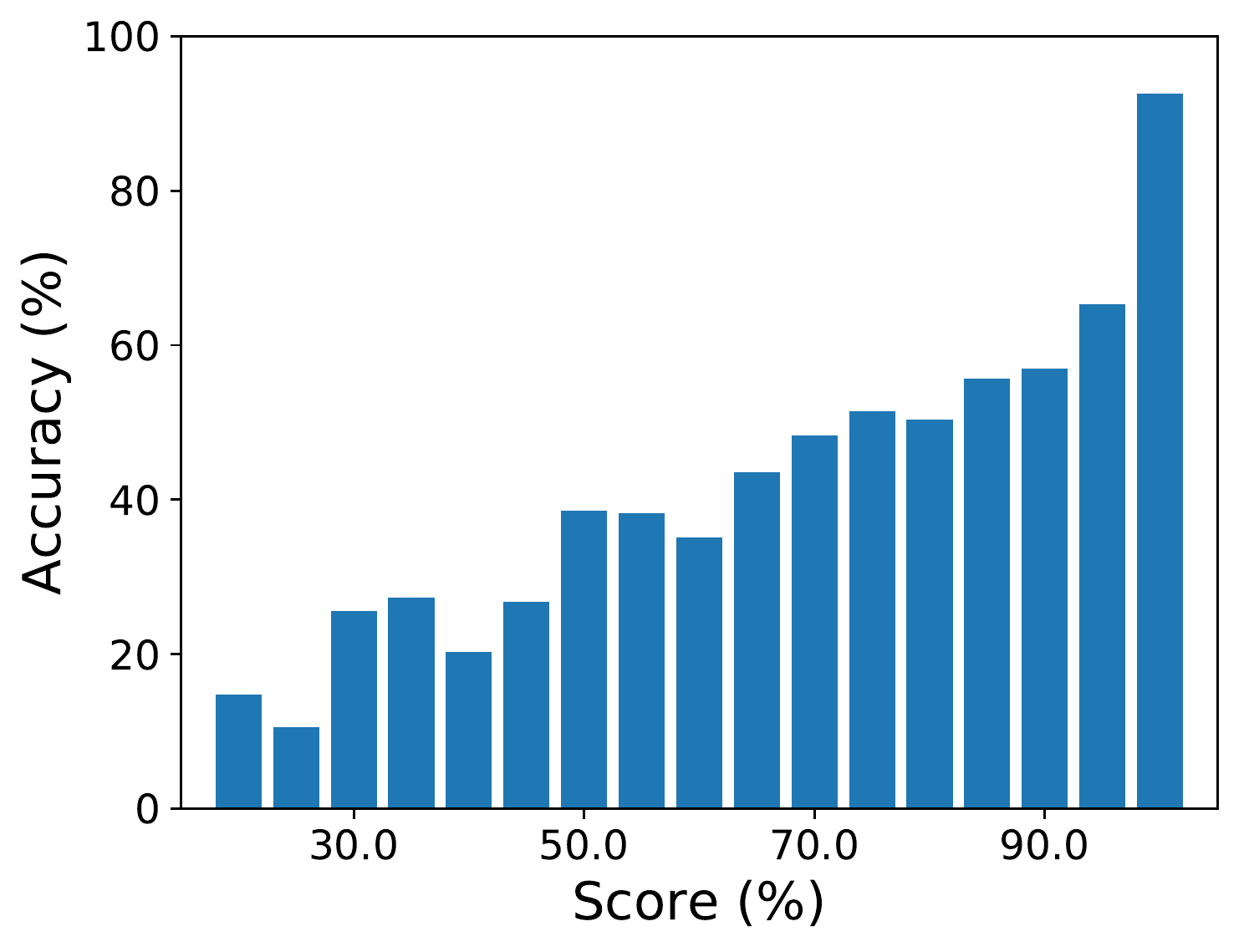}} 
    \subfigure[Prediction uncertainty characteristics with dropout in VGGNet with 16 layers]{
    	\includegraphics[width=0.33\linewidth]{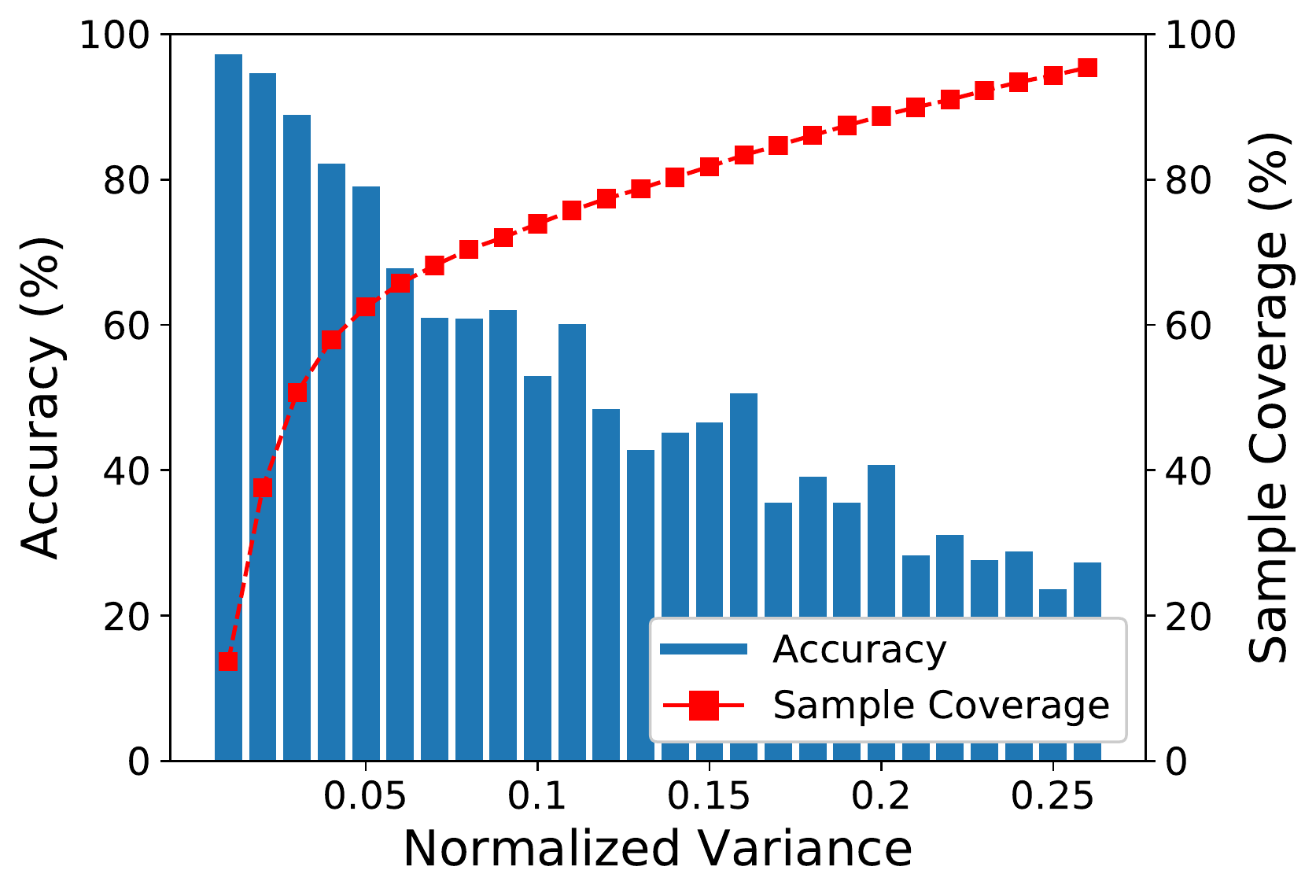}
    	\includegraphics[width=0.33\linewidth]{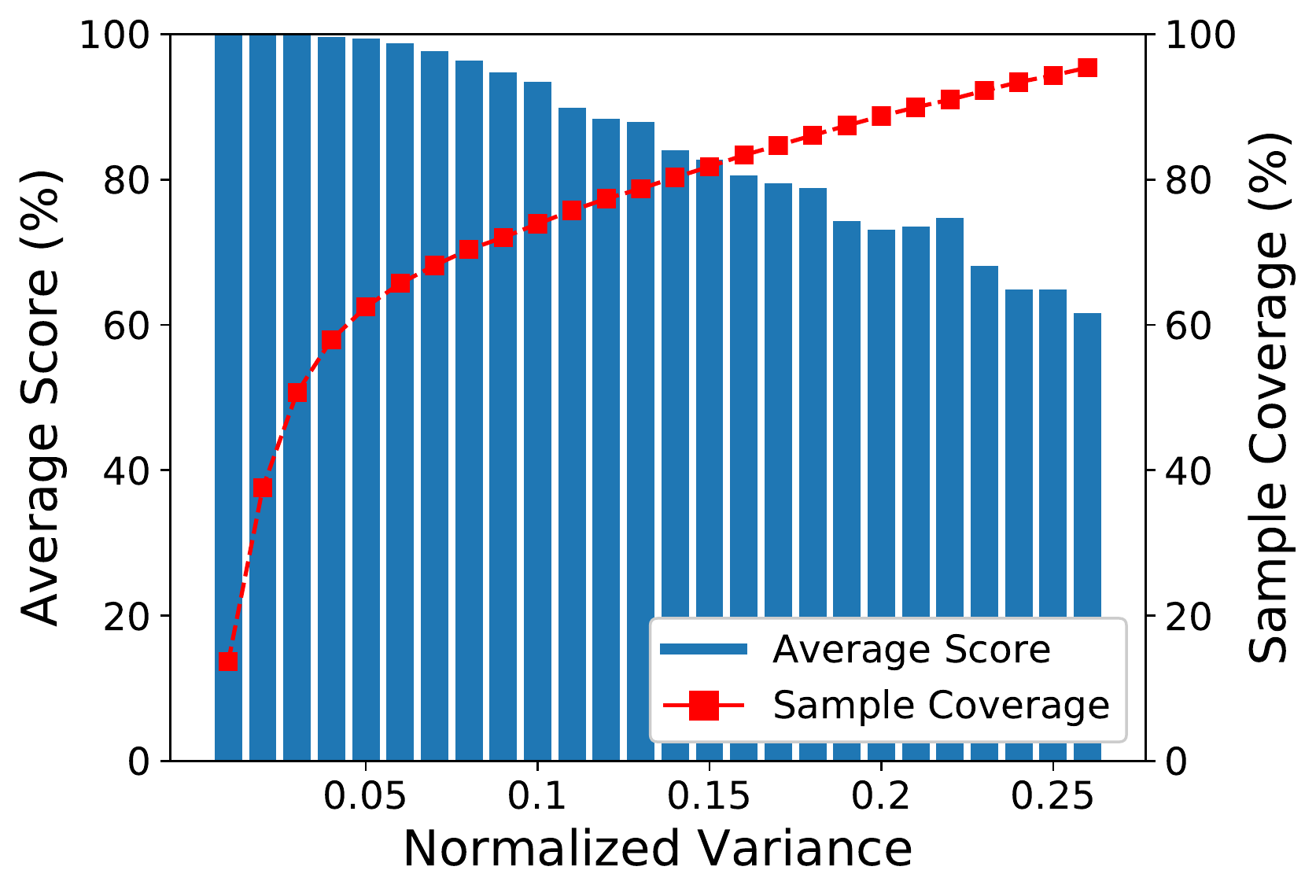}
    	\includegraphics[width=0.295\linewidth]{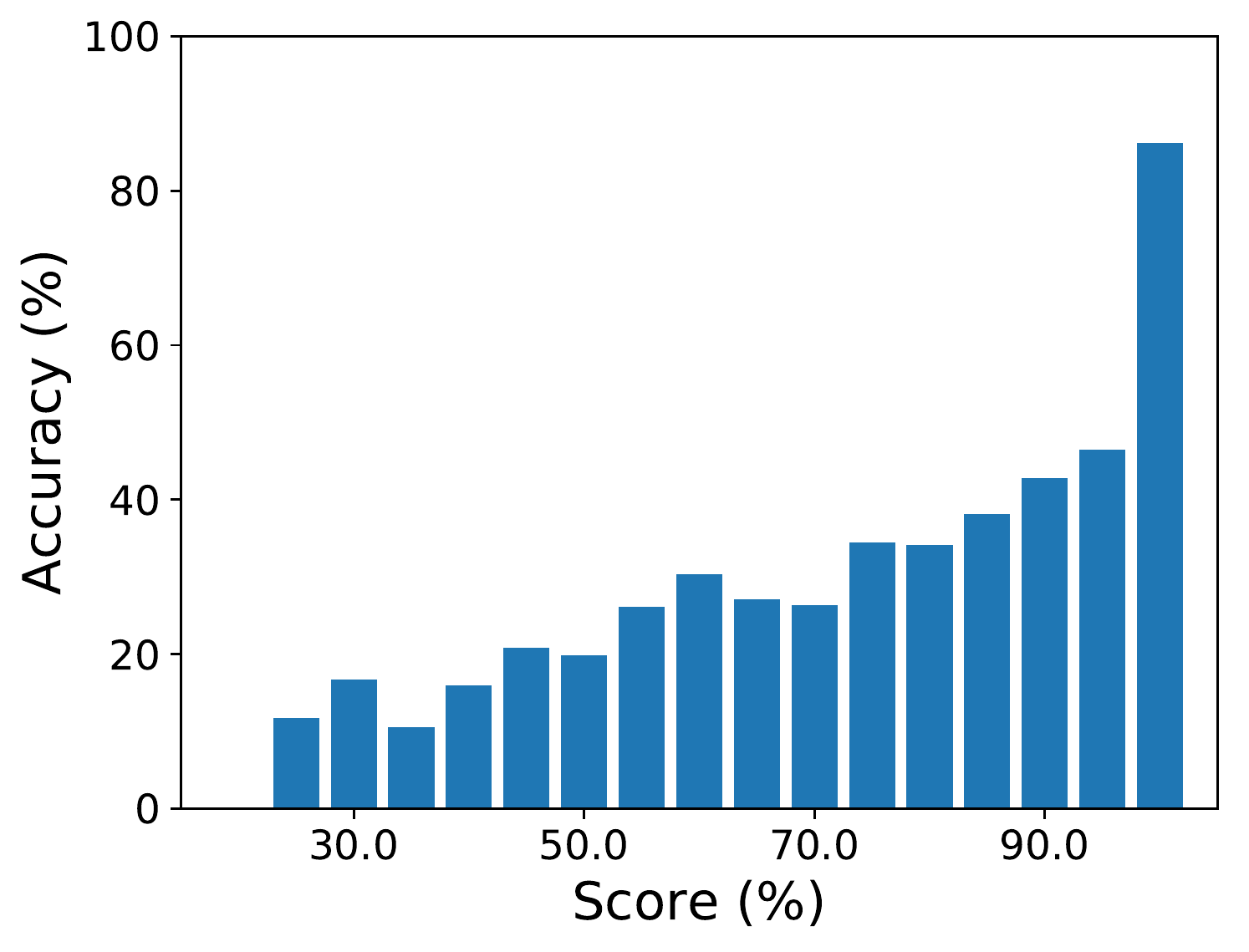}}
    \caption{Uncertainty observed from multiple stochastic inferences with two stochastic regularization methods, (a) stochastic depth and (b) dropout.  We present (left, middle) tendency of accuracy and score of the average prediction with respect to normalized variance of stochastic inferences and (right) relation between score and accuracy.  In regularization methods, average accuracy and score drop gradually as normalized variance increases.  The red lines indicate coverage (cumulative ratio) of examples.  We present results from CIFAR-100.}
    \label{fig:var_acc_hist}
    \vspace{-0.2cm}
\end{figure*}

\subsection{Empirical Observations}
\eqref{eq:variance} implies that the variation of models results in the variance of multiple stochastic predictions for a single example.
Figure~\ref{fig:var_acc_hist} presents how the variance of multiple stochastic inferences given by stochastic depth or dropout is related to the accuracy and confidence of the corresponding average prediction, where the confidence is measured by the maximum score of the average prediction. 
In the figure, the accuracy and the score of each bin are computed with the examples belonging to the corresponding bin of the normalized variance.
We present results from CIFAR-100 with ResNet-34 and VGGNet with 16 layers.
The histograms illustrate the strong correlation between the predicted variance and the reliability---accuracy and confidence---of a prediction; we can estimate accuracy and uncertainty of an example effectively based on its prediction variances given by multiple stochastic inferences.

\subsection{\hspace{-0.05cm}Variance-Weighted Confidence-Integrated Loss}
\label{sub:variance}

The strong correlation of accuracy and confidence with the predicted variance observed in Figure~\ref{fig:var_acc_hist} shows great potential to make confidence-calibrated predictions through stochastic inferences.
However, variance computation involves multiple stochastic inferences by executing multiple forward passes.
Note that this property incurs additional computational cost and may produce inconsistent results.

To alleviate these limitations, we propose a generic framework for training accuracy-score calibrated networks whose prediction score from a single forward pass directly provides the confidence of a prediction.
This objective achieved by designing a new loss function, which augments a confidence-calibration term to the standard cross-entropy loss, while the two terms are balanced by the variance measured by multiple stochastic inferences.
Specifically, our variance-weighted confidence-integrated loss $\mathcal{L}_\mathrm{VWCI}(\cdot)$ for the whole training data $(x_i,y_i) \in \mathcal{D}$ is defined by a linear combination of the standard cross-entropy loss with the ground-truth $ \mathcal{L}_\mathrm{GT}(\cdot)$ and the cross-entropy with a uniform distribution $\mathcal{L}_\mathrm{U}(\cdot)$, which is formally given by
\begin{align}
    \mathcal{L}_\mathrm{VWCI}(\theta) = &\sum_{i=1}^N (1-\alpha_i) \mathcal{L}^{(i)}_\mathrm{GT}(\theta) + \alpha_i \mathcal{L}^{(i)}_\mathrm{U}(\theta)  \nonumber \\
    = &~\frac{1}{T} \sum_{i=1}^N \sum_{j=1}^T  -(1-\alpha_i) \log p(y_i|x_i,\hat{\omega}_{i,j}) \nonumber \\
    &+ \alpha_i D_\mathrm{KL}(\mathcal{U}(y) || p(y|x_i, \hat{\omega}_{i,j})) + \xi_i   \label{eq:loss}
\end{align}
where $\alpha_i \in [0,1]$ is a normalized variance, $\hat{\omega}_{i,j} (=\theta\odot \epsilon_{i,j})$ is a sampled model parameter with binary noise for stochastic prediction, $T$ is the number of stochastic inferences, and $\xi_i$ is a constant.

The two terms in our variance-weighted confidence-integrated loss pushes the network toward the opposite directions;
the first term encourages the network to fit the ground-truth label while the second term forces the network to make a prediction close to the uniform distribution.
These terms are linearly interpolated by an instance-specific balancing coefficient $\alpha_i$, which is given by normalizing the prediction variance of an example obtained from multiple stochastic inferences.
Note that the normalized variance $\alpha_i$ is distinct for each training example and is used to measure model uncertainty.
Therefore, the optimization of our loss function produces gradient signals, which lead the predictions toward a uniform distribution for the examples with high uncertainty derived by high variances while increasing the prediction scores of the examples with low variances.

By training deep neural networks using the proposed loss function, we estimate the uncertainty of each testing example with a single forward pass.
Unlike the ordinary models, a prediction score of our model is well-calibrated and represents confidence of a prediction, which means that we can rely more on the predictions with higher scores.

\subsection{Confidence-Integrated Loss}

Our claim is that an adaptive combination of the cross-entropy losses with respect to the ground-truth and a uniform distribution is a reasonable choice to learn uncertainty.
As a special case of the proposed loss, we also present a blind version of the combination, which can be used as a baseline uncertainty estimation technique.
This baseline loss function is referred to as the confidence-integrated loss, which is given by
\begin{align}
    \mathcal{L}_\mathrm{CI}(\theta) =& ~\mathcal{L}_\mathrm{GT}(\theta) + \beta \mathcal{L}_\mathrm{U}(\theta)  \nonumber \\
    =&\sum_{i=1}^N -\log p(y_i|x_i, \theta) \nonumber \\
    &+ \beta D_\mathrm{KL}(\mathcal{U}(y) || p(y|x_i, \theta) ) + \xi,
\label{eq:baseline_loss}
\end{align}
where $p(y|x_i, \theta)$ is the predicted distribution with model parameter $\theta$ and $\xi$ is a constant.
The main idea of this loss function is to regularize with a uniform distribution by expecting the score distributions of uncertain examples to be flattened first while the distributions of confident ones remain intact, where the impact of the confidence-integrated loss term is controlled by a global hyper-parameter $\beta$.

The proposed loss function is also employed in \cite{pereyra2017regularizing} to regularize deep neural networks and improve classification accuracy.
However, \cite{pereyra2017regularizing} does not discuss confidence calibration issues while presenting marginal accuracy improvement.
On the other hand, \cite{lee2018training} discusses a similar loss function but focuses on differentiating between in-distribution and out-of-distribution examples by measuring the loss of each example using only one of the two loss terms depending on its origin.

Contrary to the existing approaches, we employ the loss function in \eqref{eq:baseline_loss} to estimate prediction confidence in deep neural networks.
Although the confidence-integrated loss makes sense intuitively, such blind selection of a hyper-parameter $\beta$ limits its generality compared to our variance-weighted confidence-integrated loss.

\subsection{Relation to Other Calibration Approaches}
There are several score calibration techniques~\citep{guo2017oncalibration, zadrozny2002transforming, naeini2015obtaining, niculescu2005predicting} by adjusting confidence scores through post-processing, among which \cite{guo2017oncalibration} presents a method to calibrate confidence of predictions by scaling logits of a network using a global temperature $\tau$.
The scaling is performed before applying the softmax function, and $\tau$ is trained with a validation dataset.
As discussed in \cite{guo2017oncalibration}, this simple technique is equivalent to maximize entropy of the output distribution $p(y_i | x_i )$.
It is also identical to minimize KL-divergence $D_\mathrm{KL} ( p(y_i | x_i ) || ~\mathcal{U}(y) )$ because
\begin{align}
    &\hspace{-0.2cm}D_\mathrm{KL}(p(y_i | x_i)||~\mathcal{U}(y)) \nonumber \\
    &= \sum_{c \in C} p(y_i^c|x_i) \log p(y_i^c|x_i) -  p(y_i^c|x_i) \log \mathcal{U}(y^c) \nonumber \\
    &= -H(p(y_i|x_i)) + \xi',
\end{align}
where $C$ is a class set and $\xi'$ is a constant.
We can formulate another confidence-integrated loss with the entropy as
\begin{align}
    \mathcal{L}'_\mathrm{CI}(\theta) =  \sum_{i=1}^N -\log p(y_i|x_i, \theta) - \gamma H(p(y_i | x_i, \theta)),
    \label{eq:entropy_loss}
\end{align}
where $\gamma$ is a constant.
\eqref{eq:entropy_loss} implies that temperature scaling in \cite{guo2017oncalibration} is closely related to our framework.


\section{Experiments}
\label{sec:experiments}

\begin{table*}[t] 
\caption{
Classification accuracy and calibration scores for several combinations of network architectures and datasets.
We compare models trained with baseline, CI and VWCI losses.
Since CI loss involves a hyper-parameter $\beta$, we present mean and standard deviation of results from models with five different $\beta$'s.
In addition, we also show results from the oracle CI loss, CI[Oracle], which are the most optimistic values out of results from all $\beta$'s in individual columns.
Note that the numbers corresponding to CI[Oracle] may come from different $\beta$'s.
Refer to the supplementary document for the full results.
}
\label{tab:results_average}
\vspace{0.1cm}
\centering
\scalebox{0.835}{
\begin{tabular}{@{}cccccccc@{}}
\toprule
\multirow{1}{*}{\begin{tabular}[c]{@{}c@{}} Dataset \end{tabular}}
& \multicolumn{1}{c}{\begin{tabular}[c]{@{}c@{}} Architecture \end{tabular}}
& \multirow{1}{*}{\begin{tabular}[c]{@{}c@{}} Method \end{tabular}}
& \multicolumn{1}{c}{\begin{tabular}[c]{@{}c@{}} Accuracy [\%] \end{tabular}}
& \multicolumn{1}{c}{\begin{tabular}[c]{@{}c@{}} ECE \end{tabular}}
& \multicolumn{1}{c}{\begin{tabular}[c]{@{}c@{}} MCE \end{tabular}} 
& \multicolumn{1}{c}{\begin{tabular}[c]{@{}c@{}} NLL \end{tabular}} 
& \multicolumn{1}{c}{\begin{tabular}[c]{@{}c@{}} Brier Score \end{tabular}} \\ 

\midrule
 \multirow{16}{*}{\begin{tabular}[c]{@{}c@{}} Tiny ImageNet \end{tabular}}

%
&  \multirow{4}{*}{ResNet-34 }    
& \multicolumn{1}{c}{Baseline } & 50.82			& 0.067			& 0.147			& 2.050		& 0.628 \\
& & \multicolumn{1}{c}{CI}   & 50.09 $\pm$ 1.08    & 0.134 $\pm$ 0.079   & 0.257 $\pm$ 0.098   & 2.270 $\pm$ 0.212   & 0.665 $\pm$ 0.037 \\
& & \multicolumn{1}{c}{VWCI}   & {\bf52.80}			& {\bf0.027}			& {\bf0.076}	& {\bf1.949}	& {\bf0.605}  \\
\cdashline{3-8} 
& & \multicolumn{1}{c}{CI[Oracle]}  & 51.45    & 0.035   & 0.171   & 2.030   & 0.620 \\

\cmidrule{2-8} 
&   \multirow{4}{*}{VGG-16 }    
& \multicolumn{1}{c}{Baseline } & 46.58			& 0.346			& 0.595			& 4.220		& 0.844 \\
& & \multicolumn{1}{c}{CI}   & 46.82 $\pm$ 0.81    & 0.226 $\pm$ 0.095   & 0.435 $\pm$ 0.107   & 3.224 $\pm$ 0.468   & 0.761 $\pm$ 0.054 \\
& & \multicolumn{1}{c}{VWCI}   &  {\bf48.03}			& {\bf0.053}			& \bf{0.142}	& {\bf2.373}	& {\bf0.659} \\
\cdashline{3-8} 
& & \multicolumn{1}{c}{CI[Oracle]}  & 47.39    & 0.122   & 0.320   & 2.812   & 0.701 \\

\cmidrule{2-8} 
&   \multirow{4}{*}{WideResNet-16-8 }    
& \multicolumn{1}{c}{Baseline } & 55.92			& 0.132			& 0.237			& 1.974		& 0.593	\\
& & \multicolumn{1}{c}{CI}   & 55.80 $\pm$ 0.44    & 0.115 $\pm$ 0.040   & 0.288 $\pm$ 0.100   & 1.980 $\pm$ 0.114   & 0.594 $\pm$ 0.017 \\
& & \multicolumn{1}{c}{VWCI}   & {\bf56.66}		& {\bf0.046}		& {\bf0.136}		& {\bf1.866}	& {\bf0.569} \\
\cdashline{3-8} 
& & \multicolumn{1}{c}{CI[Oracle]}  & 56.38    & 0.050   & 0.208   & 1.851   & 0.572 \\

\cmidrule{2-8} 
&   \multirow{4}{*}{DenseNet-40-12 }    
& \multicolumn{1}{c}{Baseline } & {42.50} & {\bf0.020} & 0.154 & {2.423} & {0.716} \\
& & \multicolumn{1}{c}{CI}   & 40.18 $\pm$ 1.68    & 0.059 $\pm$ 0.061   & 0.152 $\pm$ 0.082   & 2.606 $\pm$ 0.208   & 0.748 $\pm$ 0.035 \\
& & \multicolumn{1}{c}{VWCI}   & {\bf43.25} & {0.025} & {\bf0.089} & {\bf2.410} & {\bf0.712} \\
\cdashline{3-8} 
& & \multicolumn{1}{c}{CI[Oracle]}  & 41.21    & 0.025   & 0.094   & 2.489   & 0.726 \\

\midrule
 \multirow{16}{*}{\begin{tabular}[c]{@{}c@{}} CIFAR-100 \end{tabular}} 
%
&   \multirow{4}{*}{ResNet-34 }    
& \multicolumn{1}{c}{Baseline } & 77.19			& 0.109		& 0.304		& 1.020		& 0.345\\
& & \multicolumn{1}{c}{CI}   & 77.56 $\pm$ 0.60    & 0.134 $\pm$ 0.131   & 0.251 $\pm$ 0.128   & 1.064 $\pm$ 0.217   & 0.360 $\pm$ 0.057 \\
& & \multicolumn{1}{c}{VWCI}   & {\bf78.64}    & {\bf0.034}   & {\bf0.089}   & {\bf0.908}   & {\bf0.310} \\
\cdashline{3-8} 
& & \multicolumn{1}{c}{CI[Oracle]}  & 78.54    & 0.029   & 0.087   & 0.921   & 0.321 \\

\cmidrule{2-8} 
&   \multirow{4}{*}{VGG-16 }    
& \multicolumn{1}{c}{Baseline } & 73.78			& 0.187	& 0.486	& 1.667	& 0.437\\
& & \multicolumn{1}{c}{CI}   & 73.75 $\pm$ 0.35    & 0.183 $\pm$ 0.079   & 0.489 $\pm$ 0.214   & 1.526 $\pm$ 0.175   & 0.436 $\pm$ 0.034 \\	
& & \multicolumn{1}{c}{VWCI}   &  {\bf73.87}	&  {\bf0.098}	& {\bf0.309}	& {\bf1.277}	& {\bf0.391}  \\
\cdashline{3-8} 
& & \multicolumn{1}{c}{CI[Oracle]}  & 73.78    & 0.083   & 0.285   & 1.289   & 0.396 \\
	
\cmidrule{2-8} 
&   \multirow{4}{*}{WideResNet-16-8 }    
& \multicolumn{1}{c}{Baseline } & 77.52			& 0.103			& 0.278				& 0.984			& 0.336	\\
& & \multicolumn{1}{c}{CI}  	 & 77.35 $\pm$ 0.21    & 0.133 $\pm$ 0.091   	& 0.297 $\pm$ 0.108   	& 1.062 $\pm$ 0.180   & 0.356 $\pm$ 0.044 \\
& & \multicolumn{1}{c}{VWCI}   &{\bf 77.74}    & {\bf0.038}   & {\bf0.101}   & {\bf0.891}   & {\bf0.314} \\
\cdashline{3-8} 
& & \multicolumn{1}{c}{CI[Oracle]}  & 77.53    & 0.074   & 0.211   & 0.931   & 0.327 \\

\cmidrule{2-8} 
&   \multirow{4}{*}{DenseNet-40-12 }    
& \multicolumn{1}{c}{Baseline } & 65.91    & 0.074   					& 0.134   & 1.238   & 0.463 \\
& & \multicolumn{1}{c}{CI}   & 64.72 $\pm$ 1.46    & 0.070 $\pm$ 0.040   & 0.138 $\pm$ 0.055   & 1.312 $\pm$ 0.125   & 0.482 $\pm$ 0.028 \\
& & \multicolumn{1}{c}{VWCI}   & {\bf67.45}    & {\bf0.026}   			& {\bf0.094}   & {\bf1.161}   & {\bf0.439} \\
\cdashline{3-8} 
& & \multicolumn{1}{c}{CI[Oracle]}  & 66.20    & 0.019   & 0.053   & 1.206   & 0.456 \\
\bottomrule
\end{tabular}}
\vspace{-0.2cm}
\end{table*}

\subsection{Experimental Settings}

We select four most widely used deep neural network architectures to test the proposed algorithm: ResNet~\citep{he2016deep}, VGGNet~\citep{simonyan2015very}, WideResNet~\citep{zagoruyko2016wide} and DenseNet~\citep{huang2017dense}.

We employ stochastic depth in ResNet as proposed in \cite{he2016deep} while employing dropouts~\citep{srivastava2014dropout} before every {\sf fc} layer except for the classification layer in other architectures.
Note that, as discussed in Section~\ref{sec:sr_as_bayesian}, both stochastic depth and dropout inject multiplicative binary noise to within-layer activations or residual blocks, they are equivalent to noise injection into network weights.
Hence, training with $\ell_2$ regularization term enables us to interpret stochastic depth and dropout by Bayesian models.

We evaluate the proposed framework on two benchmarks, 
Tiny ImageNet and CIFAR-100, which contain $64\times64$ images in 200 object classes and $32\times32$ images in 100 object classes, respectively.
There are 500 training images per class in both datasets. 
For testing, we use the validation set of Tiny ImageNet and the test set of CIFAR-100, which have 50 and 100 images per class, respectively.
To test the two benchmarks with the same architecture, we resize images in Tiny ImageNet to $32\times32$.

All networks are trained by a stochastic gradient decent method with the momentum 0.9 for 300 epochs.
We set the initial learning rate to 0.1 with the exponential decay in a factor of 0.2 at epoch 60, 120, 160, 200 and 250.
Each batch consists of 64 training examples for ResNet, WideResNet and DenseNet and 256 for VGGNet.
To train networks with the proposed variance-weighted confidence-integrated loss, we draw $T$ samples with network parameters $\omega_i$ for each input image, and compute the normalized variance $\alpha$ based on $T$ forward passes.
The normalized variance is given by the mean of the Bhattacharyya coefficients between individual predictions and the average prediction, and, consequently, in the range of $[0.1]$.

\begin{table*}[h] 
\caption{
Comparison between VWCI and TS of multiple datasets and architectures.}
\label{tab:ts}
\vspace{0.1cm}
\centering
\scalebox{0.835}{
\begin{tabular}{@{}cccccccc@{}}
\toprule
\multirow{1}{*}{\begin{tabular}[c]{@{}c@{}} Dataset \end{tabular}}
& \multicolumn{1}{c}{\begin{tabular}[c]{@{}c@{}} Architecture \end{tabular}}
& \multirow{1}{*}{\begin{tabular}[c]{@{}c@{}} Method \end{tabular}}
& \multicolumn{1}{c}{\begin{tabular}[c]{@{}c@{}} Accuracy [\%] \end{tabular}}
& \multicolumn{1}{c}{\begin{tabular}[c]{@{}c@{}} ECE \end{tabular}}
& \multicolumn{1}{c}{\begin{tabular}[c]{@{}c@{}} MCE \end{tabular}} 
& \multicolumn{1}{c}{\begin{tabular}[c]{@{}c@{}} NLL \end{tabular}} 
& \multicolumn{1}{c}{\begin{tabular}[c]{@{}c@{}} Brier Score \end{tabular}} \\

\midrule
 \multirow{12}{*}{\begin{tabular}[c]{@{}c@{}} Tiny ImageNet \end{tabular}}

&   \multirow{3}{*}{ResNet-34 }    
& \multicolumn{1}{c}{TS (case 1)}   & 50.82			& 0.162			& 0.272			& 2.241		& 0.660			\\
&& \multicolumn{1}{c}{TS (case 2)}   & 47.20			& {\bf0.021}			& 0.080			& 2.159		& 0.661			\\
&& \multicolumn{1}{c}{VWCI} & {\bf52.80}			& 0.027			& {\bf0.076}	& {\bf1.949}	& {\bf0.605}		\\

\cmidrule{2-8}
&   \multirow{3}{*}{VGG-16 }    
& \multicolumn{1}{c}{TS (case 1)}   & 46.58			& 0.358			& 0.604			& 4.425		& 0.855	\\
&& \multicolumn{1}{c}{TS (case 2)}   & 46.53			& {\bf0.028}			& {\bf0.067}			& {\bf2.361}		& 0.671	\\
&& \multicolumn{1}{c}{VWCI} & {\bf48.03}			& 0.053			& 0.142	& 2.373	& {\bf0.659}	\\

\cmidrule{2-8}
&   \multirow{3}{*}{WideResNet-16-8}    
& \multicolumn{1}{c}{TS (case 1)}   & 55.92			& 0.200			& 0.335			& 2.259		& 0.627 \\
&& \multicolumn{1}{c}{TS (case 2)}   & 53.95			& {\bf0.027}			& 0.224			& 1.925		& 0.595 \\
&& \multicolumn{1}{c}{VWCI} & {\bf56.66}		& 0.046		& {\bf0.136}		& {\bf1.866}	& {\bf0.569} \\

\cmidrule{2-8}
&   \multirow{3}{*}{DenseNet-40-12 }    
& \multicolumn{1}{c}{TS (case 1)}   & 42.50 & 0.037 & 0.456 & 2.436 & 0.717 \\
&& \multicolumn{1}{c}{TS (case 2)}  & 41.63 & {\bf0.024} & 0.109 & 2.483 & 0.728 \\ 
&& \multicolumn{1}{c}{VWCI}  & {\bf43.25} & 0.025 & {\bf0.089} & {\bf2.410} & {\bf0.712} \\ 

\midrule
 \multirow{12}{*}{\begin{tabular}[c]{@{}c@{}} CIFAR-100 \end{tabular}} 

&   \multirow{3}{*}{ResNet-34 }    
& \multicolumn{1}{c}{TS (case 1)}   & 77.67    & 0.133   & 0.356   & 1.162   & 0.354 \\
&& \multicolumn{1}{c}{TS (case 2)}   & 77.40    & {0.036}   & 0.165   & {\bf0.886}   & 0.323 \\
&& \multicolumn{1}{c}{VWCI} &  {\bf78.64}    & {\bf0.034}   & {\bf0.089}   & 0.908   & {\bf0.310} \\

\cmidrule{2-8}
&   \multirow{3}{*}{VGG-16 }    
& \multicolumn{1}{c}{TS (case 1)}   & 73.66    & 0.197   & 0.499   & 1.770   & 0.445 	\\
&& \multicolumn{1}{c}{TS (case 2)}   & 72.69    & {\bf0.031}   & {\bf0.074}   & {\bf1.193}   & {\bf0.389} \\
&& \multicolumn{1}{c}{VWCI} & {\bf73.87}    & 0.098   & 0.309   & 1.277   & 0.391 \\

\cmidrule{2-8}
&   \multirow{3}{*}{WideResNet-16-8}    
& \multicolumn{1}{c}{TS (case 1)}   & {77.52}    & 0.144   & 0.400   & 1.285   & 0.361 \\
&& \multicolumn{1}{c}{TS (case 2)}   & 76.42    & {\bf0.028}   & {\bf0.101}   & {\bf0.891}   & 0.332 \\
&& \multicolumn{1}{c}{VWCI} & {\bf77.74}    & 0.038   & {\bf0.101}   & {\bf0.891}   & {\bf0.314} \\

\cmidrule{2-8}
&   \multirow{3}{*}{DenseNet-40-12}    
& \multicolumn{1}{c}{TS (case 1)}   & 65.91    & 0.095   & 0.165   & 1.274   & 0.468 \\
&& \multicolumn{1}{c}{TS (case 2)}   & 64.96    & 0.082   & 0.163   & 1.306   & 0.481 \\
&& \multicolumn{1}{c}{VWCI} & {\bf67.45}    & {\bf0.026}   & {\bf0.094}   & {\bf1.161}   & {\bf0.439} \\

\bottomrule
\end{tabular}}
\vspace{-0.2cm}
\end{table*}

\subsection{Evaluation Metric}
We measure classification accuracy and calibration scores---expected calibration error (ECE), maximum calibration error (MCE), negative log likelihood (NLL) and Brier score---of the trained models.

Let $B_m$ be a set of indices of test examples whose prediction scores for the ground-truth labels fall into interval $(\frac{m-1}{M},\frac{m}{M}]$, where $M (=20)$ is the number of bins. 
ECE and MCE are formally defined by
\begin{align}
	\mathrm{ECE} &= \sum_{m=1}^{M}\frac{|B_m|}{N'}\left|\mathrm{acc}(B_m) - \mathrm{conf}(B_m)\right|, \nonumber \\
	\mathrm{MCE} &= \max_{m\in\{1, ..., M\}}\left|\mathrm{acc}(B_m) - \mathrm{conf}(B_m)\right|,  \nonumber
\end{align}
where $N'$ is the number of the test samples. Also, accuracy and confidence of each bin are given by
\begin{align}
	\mathrm{acc}(B_m) &= \frac{1}{|B_m|}\sum_{i\in B_m}\mathbbm{1}(\hat{y}_i=y_i), \nonumber \\
	\mathrm{conf}(B_m)&=\frac{1}{|B_m|}\sum_{i\in B_m}p_i \nonumber,
\end{align}
where $\mathbbm{1}$ is an indicator function, $\hat{y}_i$ and $y_i$ are predicted and true label of the $i^\text{th}$ example and $p_i$ is its predicted confidence. 
NLL and Brier score are another ways to measure the calibration \cite{hastie01statisticallearning, brier1951verification, guo2017oncalibration}, which are defined as
\begin{align}
	\mathrm{NLL} &= -\sum_{i=1}^{N'} \log p(y_i|x_i, \theta), \nonumber \\
	 \mathrm{Brier} &= \sum_{i=1}^{N'} \sum_{j=1}^{C} (p(\hat{y_i}=j|x_i, \theta)-\mathbbm{1} ({y_i} = j))^2. \nonumber
\end{align}
We note that low values for all these calibration scores means that the network is well-calibrated.

\subsection{Results}
\label{sub:results}
Table~\ref{tab:results_average} presents accuracy and calibration scores for several combinations of network architectures and benchmark datasets.
The models trained with VWCI loss consistently outperform the models with CI loss, which is a special case of VWCI, and the baseline on both classification accuracy and confidence calibration performance.
We believe that the accuracy gain is partly by virtue of the stochastic regularization with multiple samples~\cite{noh2017regularizing}. 
Performance of CI is given by the average and variance from 5 different cases of $\beta (= 1, 10^{-1}, 10^{-2}, 10^{-3}, 10^{-4})$\footnote{These 5 values of $\beta$ are selected favorably to CI based on our preliminary experiment.} and CI[Oracle] denotes the most optimistic value among the 5 cases in each column. 
Note that VWCI presents outstanding results in most cases even when compared with CI[Oracle] and that performance of CI is sensitive to the choice of $\beta$'s.
These results imply that the proposed loss function balances two conflicting loss terms effectively using the variance of multiple stochastic inferences while performance of CI varies depending on hyper-parameter setting in each dataset.

We also compare the proposed framework with the state-of-the-art post-processing method, temperature scaling (TS)~\cite{guo2017oncalibration}.
The main distinction between post-processing methods and our work is the need for held-out dataset; our method allows to calibrate scores during training without additional data while \cite{guo2017oncalibration} requires held-out validation sets to calibrate scores. 
To illustrate the effectiveness of our framework, we compare our approach with TS in the following two scenarios: 1) using the entire training set for both training and calibration and 2) using 90\% of training set for training and the remaining 10\% for calibration. 
Table~\ref{tab:ts} presents that case~1 suffers from poor calibration performance and case~2 loses accuracy substantially due to training data reduction although it shows comparable calibration scores to VWCI.
Note that TS may also suffer from the binning artifacts of histograms although we do not investigate this limitation in our work.

\subsection{Discussion}

To show the effectiveness of the propose framework, we analyze the proposed algorithm with ablative experiments.

\begin{figure}[t]
\centering
\includegraphics[width=0.9\linewidth]{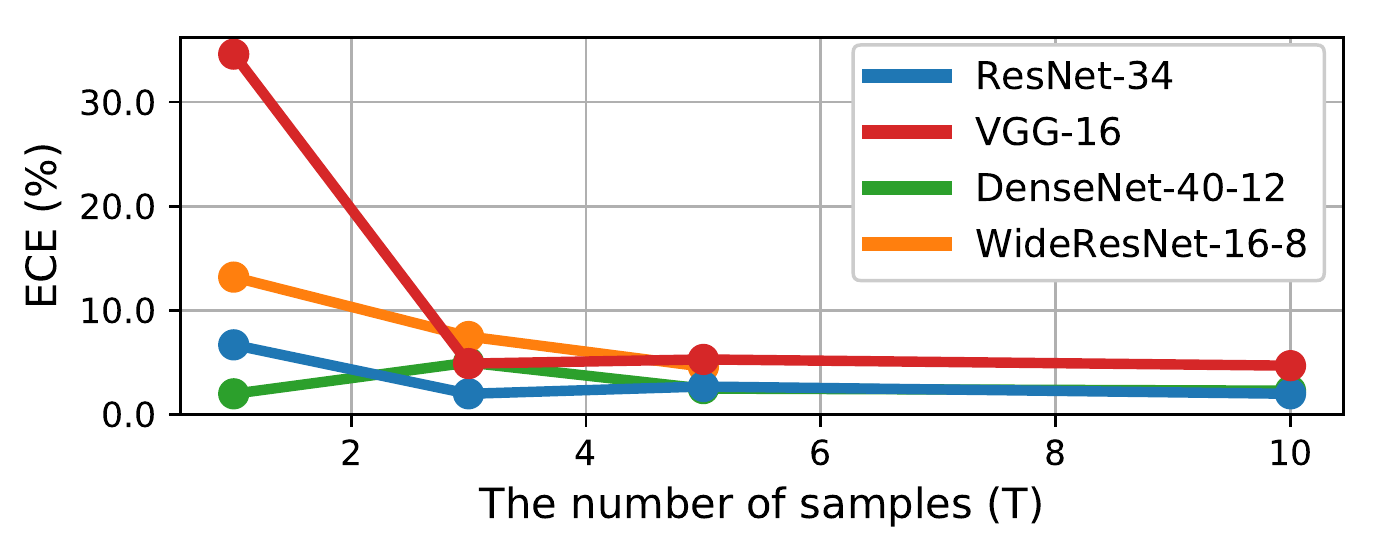}
\label{fig:b}
\includegraphics[width=0.9\linewidth]{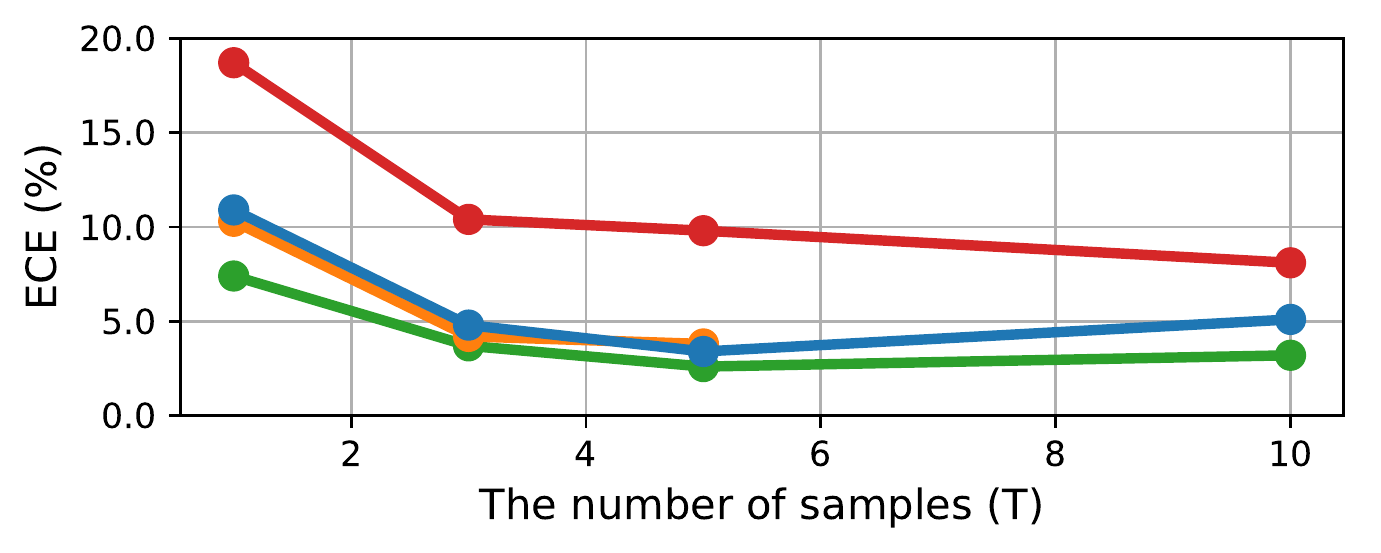}
\label{fig:a}
\vspace{-0.1cm}
\caption{ECE of VWCI loss with respect to the number of samples (T) on Tiny ImageNet (top) and CIFAR-100 (bottom) dataset. }
\label{fig:t_ece}
\end{figure}
\vspace{-0.3cm}
\paragraph{Effect of sample size for stochastic inferences} 
Figure~\ref{fig:t_ece} illustrates ECE of the models trained with our VWCI loss by varying the number of stochastic inferences  ($T$) during training.
The increase of $T$ is helpful to improve accuracy and calibration quality at the beginning but its benefit is saturated when $T$ is between 5 and 10 in general. 
Such tendency is consistent in all the tested architectures, datasets, and evaluation metrics including ECE.
We set $T$ to 5 in all the experiments.

\begin{table}[t] 
\caption{Comparisons between the models based on the VWCI losses, trained from scratch and  the uncalibrated pretrained networks (denoted by VWCI*).}
\vspace{0.1cm}
\label{table:finetuning}
\centering
\scalebox{0.835}{
\setlength\tabcolsep{4pt} \hspace{-0.3cm}

\begin{tabular}{@{}cccccccc@{}}

\toprule
\multirow{1}{*}{\begin{tabular}[c]{@{}c@{}}  \end{tabular}}
&\multicolumn{1}{c}{\begin{tabular}[c]{@{}c@{}} Architecture \end{tabular}}
& \multirow{1}{*}{\begin{tabular}[c]{@{}c@{}} Method \end{tabular}}
& \multicolumn{1}{c}{\begin{tabular}[c]{@{}c@{}} Acc. [\%] \end{tabular}}
& \multicolumn{1}{c}{\begin{tabular}[c]{@{}c@{}} ECE \end{tabular}}
& \multicolumn{1}{c}{\begin{tabular}[c]{@{}c@{}} MCE \end{tabular}} 
& \multicolumn{1}{c}{\begin{tabular}[c]{@{}c@{}} NLL \end{tabular}} 
& \multicolumn{1}{c}{\begin{tabular}[c]{@{}c@{}} Brier \end{tabular}} \\

\midrule
&  \multirow{3}{*}{\begin{tabular}[c]{@{}c@{}} ResNet-34 \end{tabular}}
&{Baseline } & 77.19	& 0.109	& 0.304	& 1.020	& 0.345\\
&&{VWCI}   & {\bf{78.64}}    & {0.034}   & {0.089}   & {\bf0.908}   & {\bf0.310} \\
\cdashline{3-8} 
&&{VWCI$^*$} &77.87 & {\bf0.026} & {\bf0.069} & 1.013 & 0.346  \\

  \cmidrule{2-8}
 &\multirow{3}{*}{\begin{tabular}[c]{@{}c@{}} VGG-16 \end{tabular}}
&{Baseline } & 73.78	& 0.187	& 0.486	& 1.667	& 0.437\\
&& {VWCI}   &  {73.87}	&  {0.098}	& {0.309}	& {1.277}	& {0.391}  \\
\cdashline{3-8}
\multirow{-6}{*} {\rotatebox[origin=c]{90}{Tiny ImageNet}} &&{VWCI$^*$}   &{\bf74.17}&{\bf0.074}&{\bf0.243}&{\bf1.227}&{\bf0.385}\\

\midrule
& \multirow{3}{*}{\begin{tabular}[c]{@{}c@{}} ResNet-34 \end{tabular}}
& {Baseline} & 50.82	& 0.067	& 0.147	& 2.050	& 0.628 \\
&&{VWCI}   & {\bf52.80}	& \bf0.027	& {\bf0.076}& {\bf1.949}	& {\bf0.605}	\\
\cdashline{3-8}
 &&{VWCI$^*$}   & 52.77 & 0.034 & 0.099 & 1.965 & \bf0.605\\
  
  \cmidrule{2-8}
 &\multirow{3}{*}{\begin{tabular}[c]{@{}c@{}} VGG-16 \end{tabular}}
& {Baseline} & 46.58	& 0.346	& 0.595	& 4.220	& 0.844 \\
&&{VWCI}   & {\bf48.03}	& \bf0.053	& \bf0.142	& \bf2.373	& {\bf0.659}	\\
\cdashline{3-8}
\multirow{-6}{*} {\rotatebox[origin=c]{90}{CIFAR-100}} && {VWCI$^*$} &46.98 & {0.056} & {0.162} & 2.446 & 0.683  \\

\bottomrule
\end{tabular}}
\vspace{-0.3cm}
\end{table}
\vspace{-0.3cm}
\paragraph{Training cost}  
Although our approach allows single-shot confidence calibration at test-time, it increases time complexity for training due to multiple stochastic inferences. 
Fortunately, the calibrated models can be trained more efficiently without stochastic inferences in the majority ($\geq 80\%$) of iterations by initializing the networks with the pretrained baseline models.
Table~\ref{table:finetuning} confirms that the performance of our models trained from the uncalibrated pretrained models is as competitive as (or often even better than) the ones trained from scratch with the VWCI losses.

\begin{figure}[t]
    \centering
    	\includegraphics[width=0.9\linewidth]{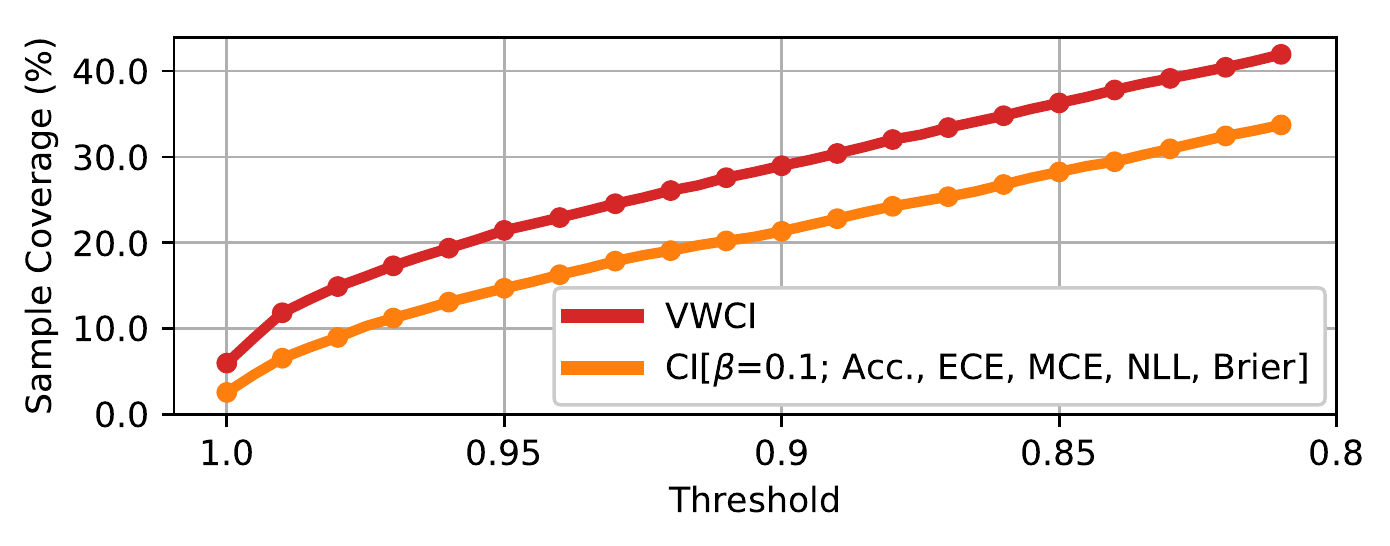}
   	\includegraphics[width=0.9\linewidth]{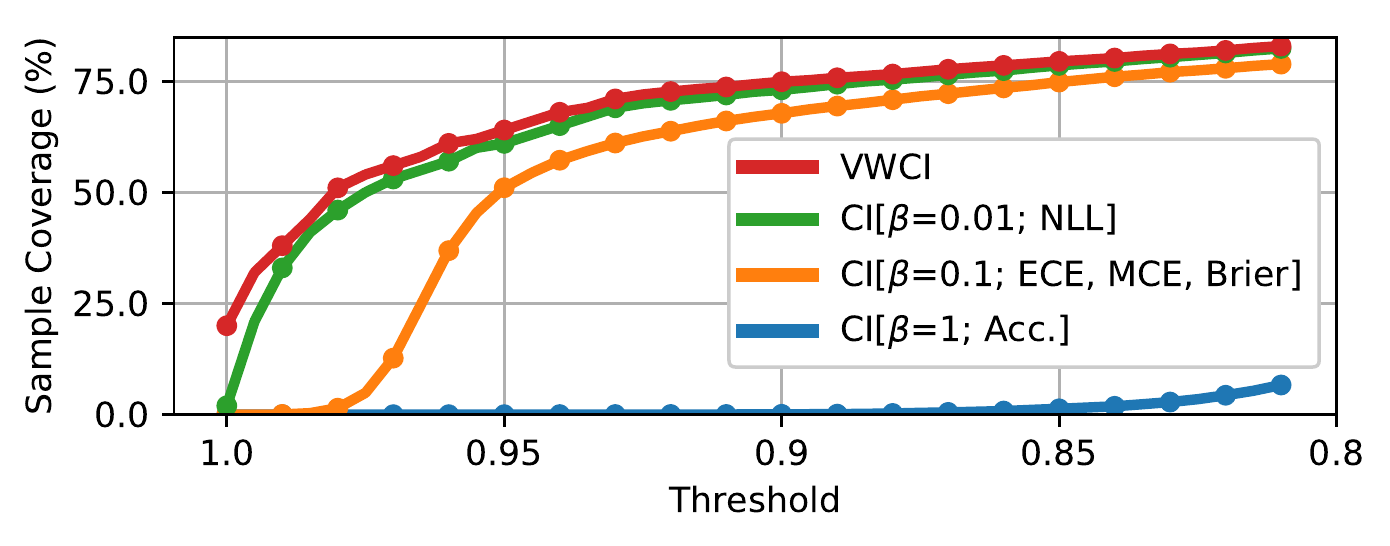}
	\vspace{-0.1cm}
    \caption{Coverage of ResNet-34 models with respect to confidence interval on Tiny ImageNet (top) and CIFAR-100~(bottom).
    The coverage is computed by the portion of examples with higher accuracy and confidence than the thresholds shown in $x$-axis.
   We present results from multiple CI models with best performances with respect to individual metrics, which are shown in the legends.
    }
    \label{fig:conf_interval}
    \vspace{-0.3cm}
\end{figure}
\vspace{-0.3cm}
\paragraph{Reliability} 
Our approach effectively maintains examples with high accuracy and confidence, which is a desirable property for building reliable real-world systems.
Figure~\ref{fig:conf_interval} illustrates portion of test examples with higher accuracy and confidence than the various thresholds in ResNet-34, where VWCI presents better coverage of the examples than CI[Oracle].
Note that coverage of CI often depends on the choice of $\beta$ significantly as demonstrated in Figure~\ref{fig:conf_interval} (right) while VWCI maintains higher coverage than CI using accurately calibrated prediction scores.
These results imply that using the predictive uncertainty for balancing the loss terms is preferable to setting with a constant coefficient.

\section{Conclusion}
\label{sub:conclusion}
We presented a generic framework for uncertainty estimation of a prediction in deep neural networks by calibrating accuracy and score based on stochastic inferences.
Based on Bayesian interpretation of stochastic regularization and our empirical observation results, we claim that variation of multiple stochastic inferences for a single example is a crucial factor to estimate uncertainty of the average prediction.
Inspired by this fact, we design the variance-weighted confidence-integrated loss to learn confidence-calibrated networks and enable uncertainty to be estimated by a single prediction.
The proposed algorithm is also useful to understand existing confidence calibration methods in a unified way, and we compared our algorithm with other variations within our framework to analyze their properties.


\vspace{-0.3cm}
\paragraph{Acknowledgments} 
This work was partly supported by Samsung Advanced Institute of Technology and Korean ICT R\&D program of the MSIP/IITP grant [2014-0-00059, 2017-0-01778].

{\small
\bibliographystyle{ieee_fullname}
\bibliography{cvpr2019_conference}
}

\onecolumn

\setcounter{section}{0}
\renewcommand\thesection{\Alph{section}}

\section{Stochastic depth as approximate Bayesian inference}
\label{appendix:stochastic}

ResNet \citep{he2016deep} proposes to add skip connections to the network. If $x_i$ denotes the output of the $i^{th}$ layer and $f_i(x)$ represents a typical convolutional transformation, we can obtain the forward propagation
\begin {align}
\label{eq:resnet}
	\rvx_i = f_i(\rvx_{i-1}) + \rvx_{i-1}
\end {align}
and $f_i(\rvx)$ is commonly defined by
\begin {align}
	f_i(\rvx) = W_i \cdot \sigma(B({W_i}' \cdot \sigma(B(\rvx))))
\end {align}

where $W_i$ and ${W_i}'$ are weight matrices, ($\cdot$) denotes convolution, and $B$ and $\sigma$ indicates batch normalization and ReLU function, respectively.

ResNet with stochastic depth \citep{huang2016deep} randomly drops a subset of layers and bypass them with short-cut connection. Let $e_i \in \{0, 1\}$ denotes a Bernoulli random variable which indicates whether the $i^{th}$ residual block is active or not. The forward propagation is extended from \Eqref{eq:resnet} to 
\begin {align}
	\rvx_i = e_i f_i(\rvx_{i-1}) + \rvx_{i-1}.
\end {align}
Now we can transform stochasticity from layers to the parameter space as follows:
\begin {align}
	e_i f_i (\rvx_{i-1}) + \rvx_{i-1} &= e_i \big ( W_i \cdot \sigma(B({W_i}' \cdot \sigma(B'(\rvx_{i-1})))) \big)+ \rvx_{i-1} \\
	&= e_i^4 \big ( W_i \cdot \sigma(B({W_i}' \cdot \sigma(B'(\rvx_{i-1})))) \big)+ \rvx_{i-1} \\
	&= e_i^4 \bigg (W_i \cdot \sigma(\gamma_i({{{W_i}' \cdot \sigma(B'(\rvx_{i-1}))-\mu_i}\over{\delta_i}})+\beta_i ) \bigg)+ \rvx_{i-1} \\
	&= e_i W_i \cdot \sigma(
	e_i \begin{bmatrix}
	\gamma_i \\ \beta_i
	\end{bmatrix}^T
	\begin{bmatrix}
	{1\over{\delta_i}}({{{e_i W_i}' \cdot \sigma(e_i \gamma_i'(\overline{\rvx_{i-1}}+\beta_i'))-e_i \mu_i}}) \\ 1  
	\end{bmatrix} ) + \rvx_{i-1}\\
	&= \widetilde{W_i} \cdot \sigma(
	\begin{bmatrix}
	\widetilde{\gamma_i} \\ \widetilde{\beta_i}
	\end{bmatrix}^T
	\begin{bmatrix}
	{1\over{\delta_i}}({{{\widetilde{W_i}}' \cdot \sigma(\widetilde{\gamma_i'}(\overline{\rvx_{i-1}}+\widetilde{\beta_i'}))-e_i \mu_i}}) \\ 1  
	\end{bmatrix} ) + \rvx_{i-1}\\
	&= f_{\widetilde{W_i}, \widetilde{W_i}', \widetilde{\gamma_i}, \widetilde{\gamma_i}', \widetilde{\beta_i}, \widetilde{\beta_i}' } (\rvx_{i-1}) = f_{\widetilde{\omega_i}} (\rvx_{i-1})
\end {align} 
$e_i = e_i^4$ since it is a Bernoulli random variable. All stochastic parameters ${\widetilde{\omega}_i} = {[\widetilde{W_i}, \widetilde{W_i}', \widetilde{\gamma_i}, \widetilde{\gamma_i}', \widetilde{\beta_i}, \widetilde{\beta_i}' ]}$ in this block drop at once or not.

\clearpage

\section{Approximation of KL-Divergence}
\label{appendix:approximation}

Let  $\omega \in \mathbb{R}^D$, $p(\omega) = \mathcal{N}(0,\mathbf{I})$ and $q_\theta(\omega) = \sum_{i=1}^2 e_i \mathcal{N}(\theta_i, \sigma^2\mathbf{I})$ with a probability vector $\mathbf{e} = (e_1, e_2)$ where $e_i \in [0, 1]$ and $ \sum_{i=1}^2 e_i=1$. In our work, $\theta_1$ denotes the deterministic model parameter $[{W_i}, {W_i}', {\gamma_i}, {\gamma_i}', {\beta_i}, {\beta_i}' ]$ and $\theta_2=0$. The KL-Divergence between $q_\theta(\omega)$ and $p(\omega)$ is
\begin{align}
\mathrm{KL}(q_\theta(\omega) || p(\omega)) &= \int q_\theta(\omega) \log \frac{q_\theta(\omega)}{p(\omega)}  d\omega \\
&= \int q_\theta(\omega) \log q_\theta(\omega)d\omega - \int q_\theta(\omega) \log p(\omega)  d\omega \label{eq:kl}.
\end{align}
We can re-parameterize the first entropy term with $\omega = \theta_i + \sigma \epsilon_i$ where $\epsilon_i \sim \mathcal{N}(0,\mathbf{I})$.
\begin{align}
\int q_\theta(\omega) \log q_\theta(\omega)d\omega &= \sum_{i=1}^2 e_i \int \mathcal{N}(\omega; \theta_i, \sigma^2_i\mathbf{I}) \log q_\theta(\omega) d\omega \\
 &= \sum_{i=1}^2 e_i \int \mathcal{N}(\epsilon_i; 0, \mathbf{I}) \log q_\theta(\theta_i+ \sigma \epsilon_i) d\omega
\end{align}
Using $q_\theta(\theta_i+ \sigma \epsilon_i) \approx e_i (2 \pi)^{-D/2} \sigma^{-1} \exp ( -\frac{1}{2} \epsilon_i^{T} \epsilon_i )$ for large enough $D$,
\begin{align}
\int q_\theta(\omega) \log q_\theta(\omega)d\omega &\approx \sum_{i=1}^2 e_i \int \mathcal{N}(\epsilon_i; 0, \mathbf{I}) \log\big(e_i (2 \pi)^{-D/2} \sigma^{-1} \exp ( -\frac{1}{2} \epsilon_i^{T} \epsilon_i )\big) d\omega \\
&= \sum_{i=1}^2 \frac{e_i}{2} \bigg(\log e_i - \log \sigma + \int \mathcal{N}(\epsilon_i; 0, \rmI) \epsilon_{i}^T \epsilon_{i} d \epsilon_i - D \log 2 \pi \bigg) \\
&\approx \sum_{i=1}^2 \frac{e_i}{2} \big( -\log \sigma - D (1+\log 2 \pi) \big) - \frac{1}{2}H(e).
\end{align}
For the second term of the \Eqref{eq:kl},
\begin{align}
\int q_\theta(\omega) \log p(\omega) d\omega &= \sum_{i=1}^2 e_i \int \mathcal{N}(\omega; \theta_i, \sigma^2_i\mathbf{I}) \log \mathcal{N}(\omega; 0, \mathbf{I}) d\omega \\
&= -\frac{1}{2}\sum_{i=1}^2 e_i \big( \theta_i^T \theta_i + D\sigma \big).
\end{align}
Then we can approximate
\begin{align}
\mathrm{KL}(q_\theta(\omega) || p(\omega)) &\approx \sum_{i=1}^2 \frac{e_i}{2} \big(\theta_i^T \theta_i +D\sigma - \log \sigma - D (1+\log 2 \pi) \big) - \frac{1}{2}H(e).
\end{align}
For a more general proof, see \cite{gal2016dropout}.

\clearpage
\section{Full Experimental Results}
\label{appendix:full}

We present the full results of Table~1 in the main paper including all scores from individual CI models with different $\beta$'s.
Table~\ref{tab:tinyimage_results_full} and \ref{tab:cifar100_results_full} show the results on Tiny ImageNet and CIFAR-100, respectively.

\begin{table*}[h] 
\renewcommand\thetable{A}
\caption{
Classification accuracy and calibration scores of the models trained with various architectures on Tiny ImageNet supplementing Table~1 in the main paper. 
This table includes results from all models trained with CI loss with different $\beta$'s.
In addition to the architectures presented in Table~1, we also include results from ResNet-18.
}
\label{tab:tinyimage_results_full}
\vspace{6px}
\centering
\resizebox{0.9\columnwidth}{!}{
\scalebox{0.1}{
\begin{tabular}{@{}cccccccc@{}}
\toprule
\multirow{1}{*}{\begin{tabular}[c]{@{}c@{}} Dataset \end{tabular}}
& \multicolumn{1}{c}{\begin{tabular}[c]{@{}c@{}} Architecture \end{tabular}}
& \multirow{1}{*}{\begin{tabular}[c]{@{}c@{}} Method \end{tabular}}
& \multicolumn{1}{c}{\begin{tabular}[c]{@{}c@{}} Accuracy[\%] \end{tabular}}
& \multicolumn{1}{c}{\begin{tabular}[c]{@{}c@{}} ECE \end{tabular}}
& \multicolumn{1}{c}{\begin{tabular}[c]{@{}c@{}} MCE \end{tabular}} 
& \multicolumn{1}{c}{\begin{tabular}[c]{@{}c@{}} NLL \end{tabular}} 
& \multicolumn{1}{c}{\begin{tabular}[c]{@{}c@{}} Brier Score \end{tabular}} \\ 

\midrule
 \multirow{35}{*}{\begin{tabular}[c]{@{}c@{}} Tiny ImageNet \end{tabular}} 
&   \multirow{7}{*}{ResNet-18 }    
& \multicolumn{1}{c}{Baseline} & 46.38    & 0.029   & 0.086   & 2.227   & 0.674 \\
& & \multicolumn{1}{c}{CI[$\beta=10^{-4}$]}  &46.48    & {\color{blue}0.022}   & 0.073   & 2.216   & 0.672 \\
& & \multicolumn{1}{c}{CI[$\beta=10^{-3}$]}  &47.20    & {\color{blue}0.022}   & {\color{blue}0.060}   & 2.198   & {\color{blue}0.666} \\
& & \multicolumn{1}{c}{CI[$\beta=0.01$]}  & 47.03    & {\bf0.021}   & 0.157   & {\color{blue}2.193}   & 0.667 \\
& & \multicolumn{1}{c}{CI[$\beta=0.1$]}   & 47.58    & 0.055   & 0.111   & 2.212   & {\color{blue}0.666} \\
& & \multicolumn{1}{c}{CI[$\beta=1$]}   & {\color{blue}47.92}    & 0.241   & 0.380   & 2.664   & 0.742 \\
& & \multicolumn{1}{c}{VWCI}   &{\bf48.57}    & 0.026   & {\bf0.054}   & {\bf2.129}   & {\bf0.651} \\

\cmidrule{2-8} 
&   \multirow{7}{*}{ResNet-34 }    
& \multicolumn{1}{c}{Baseline} & 50.82			& 0.067			& 0.147			& 2.050		& 0.628 \\
& & \multicolumn{1}{c}{CI[$\beta=10^{-4}$]}  & 48.89			& 0.132			& 0.241	 		& 2.257		& 0.668 \\
& & \multicolumn{1}{c}{CI[$\beta=10^{-3}$]}  & 50.17			& 0.127			& 0.227			& 2.225		& 0.653 \\
& & \multicolumn{1}{c}{CI[$\beta=0.01$]}  & 49.16			& 0.119			& 0.219			& 2.223		& 0.663 \\
& & \multicolumn{1}{c}{CI[$\beta=0.1$]}   & {\color{blue}51.45}	& {\color{blue}0.035}	& {\color{blue}0.171}	& {\color{blue}2.030}	& {\color{blue}0.620} \\
& & \multicolumn{1}{c}{CI[$\beta=1$]}   & 50.77			& 0.255			& 0.426			& 2.614			& 0.722 \\
& & \multicolumn{1}{c}{VWCI}   & {\bf52.80}			& {\bf0.027}			& {\bf0.076}	& {\bf1.949}	& {\bf0.605}  \\

\cmidrule{2-8} 
&   \multirow{7}{*}{VGG-16 }    
& \multicolumn{1}{c}{Baseline} & 46.58			& 0.346			& 0.595			& 4.220		& 0.844 \\
& & \multicolumn{1}{c}{CI[$\beta=10^{-4}$]}  & {47.26}	& 0.325			& 0.533			& 3.878		& 0.830 \\
& & \multicolumn{1}{c}{CI[$\beta=10^{-3}$]}  & {\color{blue}47.39}	& 0.296			& 0.536			& 3.542		& 0.795 \\
& & \multicolumn{1}{c}{CI[$\beta=0.01$]}  & {47.11}	& 0.259			& 0.461			& 3.046		& 0.763	 \\
& & \multicolumn{1}{c}{CI[$\beta=0.1$]}   & 46.94				& {\color{blue}0.122}	& 0.327		& {\color{blue}2.812}	& {\color{blue}0.701}	 \\
& & \multicolumn{1}{c}{CI[$\beta=1$]}   & 45.40			& 0.130			& {\color{blue}0.320} & 2.843		& 0.717 \\
& & \multicolumn{1}{c}{VWCI}   & {\bf48.03}			& {\bf0.053}			& \bf{0.142}	& {\bf2.373}	& {\bf0.659} \\

\cmidrule{2-8} 
&   \multirow{7}{*}{WideResNet-16-8 }    
& \multicolumn{1}{c}{Baseline} & 55.92			& 0.132			& 0.237			& 1.974		& 0.593	\\
& & \multicolumn{1}{c}{CI[$\beta=10^{-4}$]}  & 55.29			& 0.126			& {\color{blue}0.208}	& 1.987		& 0.598	\\ 
& & \multicolumn{1}{c}{CI[$\beta=10^{-3}$]}  & 55.53  & 0.120			& 0.237			& 1.949		& 0.592 \\
& & \multicolumn{1}{c}{CI[$\beta=0.01$]}  & 56.12			& 0.116			& 0.238			& 1.949		& 0.590 \\
& & \multicolumn{1}{c}{CI[$\beta=0.1$]}   & {\color{blue}56.38}	& {\color{blue}0.050}	& 0.456			& {\bf1.851}	& {\color{blue}0.572} \\
& & \multicolumn{1}{c}{CI[$\beta=1$]}   & 55.66			& 0.161			& 0.301			& 2.163		& 0.619 \\
& & \multicolumn{1}{c}{VWCI}   & {\bf56.66}		& {\bf0.046}		& {\bf0.136}		& {\color{blue}1.866}	& {\bf0.569} \\

\cmidrule{2-8} 
&   \multirow{7}{*}{DenseNet-40-12 }    
& \multicolumn{1}{c}{Baseline} 	& {\color{blue}42.50} & {\bf0.020} & 0.154 & {\color{blue}2.423} & {\color{blue}0.716} \\
& & \multicolumn{1}{c}{CI[$\beta=10^{-4}$]}  & 41.20 & 0.030 & 0.156 & 2.489 & 0.726 \\ 
& & \multicolumn{1}{c}{CI[$\beta=10^{-3}$]}  & 41.21 & 0.036 & 0.122 & 2.514 & 0.735 \\
& & \multicolumn{1}{c}{CI[$\beta=0.01$]}  & 40.61 & {\color{blue}0.025} & 0.097 & 2.550 & 0.739 \\
& & \multicolumn{1}{c}{CI[$\beta=0.1$]}   & 40.67 & 0.037 & {\color{blue}0.094} & 2.501 & 0.732			\\
& & \multicolumn{1}{c}{CI[$\beta=1$]}   & 37.23 & 0.169 & 0.291 & 2.975 & 0.810 \\
& & \multicolumn{1}{c}{VWCI}   & {\bf43.25} & {\color{blue}0.025} & {\bf0.089} & {\bf2.410} & {\bf0.712} \\

\bottomrule
\end{tabular}}
}
\end{table*}

\begin{table*}[t] 
\renewcommand\thetable{B}
\caption{
Classification accuracy and calibration scores of models trained with various methods on CIFAR-100 supplementing Table~1 in the main paper. 
This table includes results from all models trained with CI loss with different $\beta$'s.
In addition to the architectures presented in Table~1, we also include results from ResNet-18.
}
\label{tab:cifar100_results_full}
\vspace{6px}
\centering
\resizebox{0.9\columnwidth}{!}{
\scalebox{0.1}{
\begin{tabular}{@{}cccccccc@{}}
\toprule
\multirow{1}{*}{\begin{tabular}[c]{@{}c@{}} Dataset \end{tabular}}
& \multicolumn{1}{c}{\begin{tabular}[c]{@{}c@{}} Architecture \end{tabular}}
& \multirow{1}{*}{\begin{tabular}[c]{@{}c@{}} Method \end{tabular}}
& \multicolumn{1}{c}{\begin{tabular}[c]{@{}c@{}} Accuracy[\%] \end{tabular}}
& \multicolumn{1}{c}{\begin{tabular}[c]{@{}c@{}} ECE \end{tabular}}
& \multicolumn{1}{c}{\begin{tabular}[c]{@{}c@{}} MCE \end{tabular}} 
& \multicolumn{1}{c}{\begin{tabular}[c]{@{}c@{}} NLL \end{tabular}} 
& \multicolumn{1}{c}{\begin{tabular}[c]{@{}c@{}} Brier Score \end{tabular}} \\ 

\midrule
 \multirow{35}{*}{\begin{tabular}[c]{@{}c@{}} CIFAR-100 \end{tabular}} 
&   \multirow{7}{*}{ResNet-18 }    
& \multicolumn{1}{c}{Baseline} & 75.61			& 0.097			&0.233			& 1.024			& 0.359\\
& & \multicolumn{1}{c}{CI[$\beta=10^{-4}$]}  & 75.03		& 0.104			&0.901			&1.055			&0.369 \\
& & \multicolumn{1}{c}{CI[$\beta=10^{-3}$]}  & 75.51			& 0.087		& 0.219			& {\color{blue}0.986}			& 0.357 \\
& & \multicolumn{1}{c}{CI[$\beta=0.01$]}  & 74.95			& 0.069		& {\color{blue}0.183}			& 0.998			& 0.358 \\
& & \multicolumn{1}{c}{CI[$\beta=0.1$]}   & {\color{blue}75.94}	& {\color{blue}0.065}		& 0.961			& 1.018			& {\color{blue}0.349} \\
& & \multicolumn{1}{c}{CI[$\beta=1$]}   & 75.61			& 0.340			& 0.449			& 1.492			& 0.475 \\
& & \multicolumn{1}{c}{VWCI}   & {\bf76.09}			& {\bf0.045}			& {\bf0.128}	& {\bf0.976}	& {\bf0.342}  \\

\cmidrule{2-8} 
&   \multirow{7}{*}{ResNet-34 }    
& \multicolumn{1}{c}{Baseline} & 77.19			& 0.109		& 0.304		& 1.020		& 0.345\\
& & \multicolumn{1}{c}{CI[$\beta=10^{-4}$]}  & 77.38		& 0.105		& 0.259		& 1.000		& 0.341 \\
& & \multicolumn{1}{c}{CI[$\beta=10^{-3}$]}  & 76.98		& 0.101		& 0.261		& 0.999		& 0.344 \\
& & \multicolumn{1}{c}{CI[$\beta=0.01$]}  & 77.23		& 0.074		& 0.206		& {\color{blue}0.921}		& 0.331 \\
& & \multicolumn{1}{c}{CI[$\beta=0.1$]}   & 77.66	& {\bf0.029}	& {\bf0.087}		& 0.953		& {\color{blue}0.321} \\
& & \multicolumn{1}{c}{CI[$\beta=1$]}   & {\color{blue}78.54}			& 0.362		& 0.442		& 1.448		& 0.461 \\
& & \multicolumn{1}{c}{VWCI}   &  {\bf78.64}    &  {\color{blue}0.034}   &  {\color{blue}0.089}   & {\bf0.908}   & {\bf0.310} \\

\cmidrule{2-8} 
&   \multirow{7}{*}{VGG-16 }    
& \multicolumn{1}{c}{Baseline} & {\color{blue}73.78}			& 0.187	& 0.486	& 1.667	& 0.437\\
& & \multicolumn{1}{c}{CI[$\beta=10^{-4}$]}  & 73.19		& 0.189	& 0.860	& 1.679	& 0.446 \\
& & \multicolumn{1}{c}{CI[$\beta=10^{-3}$]}  & 73.70		& 0.183	& 0.437	& 1.585	& 0.434 \\
& & \multicolumn{1}{c}{CI[$\beta=0.01$]}  & {\color{blue}73.78}		& 0.163	& 0.425	& 1.375	& 0.420 \\
& & \multicolumn{1}{c}{CI[$\beta=0.1$]}   & 73.68	& {\bf0.083}	& {\bf0.285}	& {\color{blue}1.289}	& {\color{blue}0.396} \\
& & \multicolumn{1}{c}{CI[$\beta=1$]}   & 73.62			& 0.291	& 0.399	& 1.676	& 0.487 \\
& & \multicolumn{1}{c}{VWCI}   &  {\bf73.87}	&  {\color{blue}0.098}	& {\color{blue}0.309}	& {\bf1.277}	& {\bf0.391}  \\

\cmidrule{2-8} 
&   \multirow{7}{*}{WideResNet-16-8 }    
& \multicolumn{1}{c}{Baseline} & {77.52}			& 0.103	& 0.278	& 0.984	& 0.336	\\
& & \multicolumn{1}{c}{CI[$\beta=10^{-4}$]}  & 77.04		& 0.109	& 0.280	& 1.011	& 0.345	\\ 
& & \multicolumn{1}{c}{CI[$\beta=10^{-3}$]}  & 77.46		& 0.104	& 0.272	& 0.974	& 0.339 \\
& & \multicolumn{1}{c}{CI[$\beta=0.01$]}  & {\color{blue}77.53}		& {\color{blue}0.074}	& {\color{blue}0.211}	& {\color{blue}0.931}	& {\color{blue}0.327} \\
& & \multicolumn{1}{c}{CI[$\beta=0.1$]}   & 77.23		& 0.085	& 0.239	& 1.015	& 0.336 \\
& & \multicolumn{1}{c}{CI[$\beta=1$]}   & 77.48			& 0.295	& 0.485	& 1.378	& 0.434 \\
& & \multicolumn{1}{c}{VWCI}   & {\bf77.74}    & {\bf0.038}   & {\bf0.101}   & {\bf0.891}   & {\bf0.314} \\

\cmidrule{2-8} 
&   \multirow{7}{*}{DenseNet-40-12 }    
& \multicolumn{1}{c}{Baseline} & 65.91    & 0.074   & 0.134   & 1.238   & 0.463 \\
& & \multicolumn{1}{c}{CI[$\beta=10^{-4}$]}  & {\color{blue}66.20}    & 0.064   & 0.141   & 1.236   & 0.463  \\
& & \multicolumn{1}{c}{CI[$\beta=10^{-3}$]}  & 63.61    & 0.086   & 0.177   & 1.360   & 0.496 \\
& & \multicolumn{1}{c}{CI[$\beta=0.01$]}  & 65.13    & 0.052   & 0.127   & 1.249   & 0.471 \\
& & \multicolumn{1}{c}{CI[$\beta=0.1$]}   & 65.86    & {\bf0.019}   & {\bf0.053}   & {\color{blue}1.206}   & {\color{blue}0.456} \\
& & \multicolumn{1}{c}{CI[$\beta=1$]}   & 62.82    & 0.127   & 0.193   & 1.510   & 0.523 \\
& & \multicolumn{1}{c}{VWCI}   & {\bf67.45}    & {\color{blue}0.026}   & {\color{blue}0.094}   & {\bf1.161}   & {\bf0.439} \\

\bottomrule
\end{tabular}}
}
\end{table*}

\end{document}